\documentclass{preprint}
\usepackage[colorlinks=true]{hyperref}
\usepackage[numbers,comma,sort&compress]{natbib} 
\usepackage{graphicx}%
\usepackage{amsmath,amssymb,amsfonts}%
\usepackage{xcolor}%
\usepackage{booktabs}%
\usepackage{listings}%
\usepackage{lineno}
\usepackage{xspace}
\usepackage[most]{tcolorbox}
\usepackage{caption}
\usepackage{xtab}
\usepackage{colortbl}
\usepackage{makecell}
\usepackage{enumitem}
\usepackage{placeins}
\usepackage{cleveref}

\definecolor{dark_green}{rgb}{0, 0.5, 0}

\newcommand{\bftab}{\fontseries{b}\selectfont}

\newcommand\boxedname{Prompt\xspace} 

\definecolor{graybg}{HTML}{f1f1f1}

\newcounter{prompt}
\newenvironment{prompt}[1][]{
    \par\nolinenumbers
    \refstepcounter{prompt}
    \begin{tcolorbox}[
        breakable,
        sharp corners,
        boxrule=0.4pt,
        colback=graybg,
        colframe=black,
        coltitle=black,
        colbacktitle=graybg,
        fonttitle=\scriptsize,
        title={\textbf{\boxedname \theprompt:} #1},
        toptitle=2pt,
        bottomtitle=2pt,
        left=2pt,
        right=2pt,
        top=2pt,
        bottom=2pt,
        titlerule=0.4pt]%
    \setlength{\parindent}{0pt}%
    \setlength{\parskip}{.5em}%
    \scriptsize\ttfamily\hyphenchar\font=`\-\spaceskip=.5em plus .5em\xspaceskip=.5em%
}
{%
    \par%
    \end{tcolorbox}%
}

\lstdefinelanguage{prompt}{
    breaklines=true,
    breakindent=0pt,
    basicstyle=\scriptsize,
}

\newcommand{\ourmodel}{MonteRET\xspace}

\title{\ourmodel: AI Agent Enhancing Multimodal LLMs with Multi-granularity Knowledge Retrieval for Chest CT Report Generation}

\author[1]{Yi Lin}
\author[2]{Yihao Ding}
\author[3]{Elana Benishay}
\author[3]{Elefterios Trikantzopoulos}
\author[3]{David Nauheim}
\author[3]{Hanley Ong}
\author[4,5]{Jiang Bian}
\author[6]{Hua Xu}
\author[7]{Yuzhe Yang}
\author[3]{George Shih}
\author[1,*]{Yifan Peng}

\affil[1]{Department of Population Health Sciences, Weill Cornell Medicine, New York, USA}
\affil[2]{School of Physics, Mathematics and Computing, University of Western Australia, Crawley, Australia}
\affil[3]{Department of Radiology, Weill Cornell Medicine, New York, USA}
\affil[4]{Biostatistics and Health Data Science, School of Medicine, Indiana University, Indianapolis, USA}
\affil[5]{Regenstrief Institute, Indianapolis, USA}
\affil[6]{Department of Biomedical Informatics and Data Science, Yale School of Medicine, Yale University, New Haven, USA}
\affil[7]{Department of Computational Medicine, University of California, Los Angeles, USA}
\affil[*]{Corresponding author(s). Email(s): \url{yip4002@med.cornell.edu}}

\begin{document}
\maketitle
\let\WriteBookmarks\relax
\def\floatpagepagefraction{1}
\def\textpagefraction{.001}

\begin{abstract}
Automated chest CT report generation remains challenging because clinically faithful reporting requires both whole-volume understanding and accurate description of localized anatomical findings. Here we developed and retrospectively evaluated \ourmodel, a region-aware retrieval-enhanced framework for generating chest CT findings sections. 
\ourmodel integrates global CT features with region-level anatomical representations, retrieves clinically relevant knowledge using predicted medical conditions and region-level vision–language alignment, and refines initial reports through a knowledge-guided report rewriting agent.
We trained our model on a public cohort with 24,128 CT scans from RadGenome-ChestCT.
We evaluated \ourmodel on the public RadGenome-ChestCT test set of 1,564 CT scans and an external cohort of 82 CT scans from NewYork-Presbyterian/Weill Cornell Medical Center.
\ourmodel improved report quality, semantic similarity, and clinical efficacy compared with a matched baseline and several state-of-the-art methods. 
Gains were most pronounced for recall, suggesting fewer omitted findings. Human expert evaluation by radiology residents also favored \ourmodel. 
\end{abstract}

\section{Introduction}\label{sec1}

Pulmonary diseases, such as pulmonary embolism, pneumothorax, and pulmonary infections, remain leading contributors to global morbidity and mortality~\citep{gbd2021CausesOfDeathCollaborators2024global, who2024-hz}. 
Timely detection and reporting of thoracic abnormal findings are essential for clinical decision-making and patient management. 
Computed tomography (CT) provides high-resolution cross-sectional imaging and supports detailed assessment of thoracic disease~\citep{Wang2024-vz}, but interpretation requires considerable radiological expertise~\citep{Topol2019-pz}. 
Increasing imaging volumes and workforce constraints have motivated the development of automated systems for CT report generation~\citep{chen2025large, Liu2021-wi, Deng2025-wl, Chen2026-sd, wu2025towards, hamamci2024ct2rep, Hu2025-sj}. 

Recent advances in deep learning and vision-language modeling have enabled automated analysis of medical images across tasks such as disease classification, visual grounding, visual question answering, and report generation~\citep{Blankemeier2024-ry, bai2024m3d, Wasserthal2023-rw, Ma2024-bm, Ichinose2023-fs, zhao2025large-vocabulary, hamamci2024developing, Zhong2025-wd, Su2025-dh}. 
However, generating chest CT reports remains challenging.
First, precise CT interpretation requires both global understanding of the entire scan and detailed attention to localized anatomical regions~\citep{Patel2019-we}.
Small abnormal findings, such as a subtle nodule at the lung apex or a mild pleural effusion, may be missed if the model attends primarily to whole-image representations~\citep{Patel2019-we}.
Second, reports generated by large language models (LLMs) can contain factual inconsistencies and findings that are not grounded in medical knowledge, due to limitations in single-round generation designs and insufficient integration of external knowledge~\citep{chen2025large}.

To address these limitations, we introduce \ourmodel\footnote{\ourmodel is inspired by a poker game ``Monte'', where players combine multiple cards to form a hand. Analogously, we employ multiple knowledge-retrieval strategies to provide a richer context for the LLM. The name can also be interpreted as \textbf{M}ultim\textbf{O}dal \textbf{N}etwork for medical repor\textbf{T} generation \textbf{E}nhanced by knowledge \textbf{RET}rieval.}, a vision-language framework for chest CT report generation that combines multi-granularity anatomical modeling with knowledge-augmented retrieval. 
\ourmodel first generates a draft report by jointly encoding global volumetric features and region-level representations to capture both holistic and localized imaging findings. It then retrieves relevant knowledge from a database of prior reports using predicted medical conditions and region-level vision–language similarity. Finally, it applies a refinement agent to integrate the retrieved contextual knowledge and enhance coherence, specificity, and clinical fidelity.
By decomposing report generation into interpretable stages, \ourmodel approximates the structured and iterative review process used in radiological reporting.

We trained \ourmodel on 24,128 CT scans from the public RadGenome-ChestCT cohort and evaluated it on 1,564 held-out RadGenome-ChestCT scans and an external cohort of 82 CT scans from NewYork-Presbyterian/Weill Cornell Medical Center. 
Performance was assessed using natural language generation metrics, clinical efficacy across 18 medical conditions, anatomical region-level analyses, ablation studies, and blinded expert review. 
\ourmodel improved report quality and clinical fidelity compared with a matched baseline and existing state-of-the-art methods. Empirical results show that region-aware modeling combined with condition-guided retrieval improves clinically relevant completeness, particularly recall, while maintaining partial generalizability in an external cohort.

\begin{figure}[thbp]
    \centering
    \includegraphics[width=\textwidth]{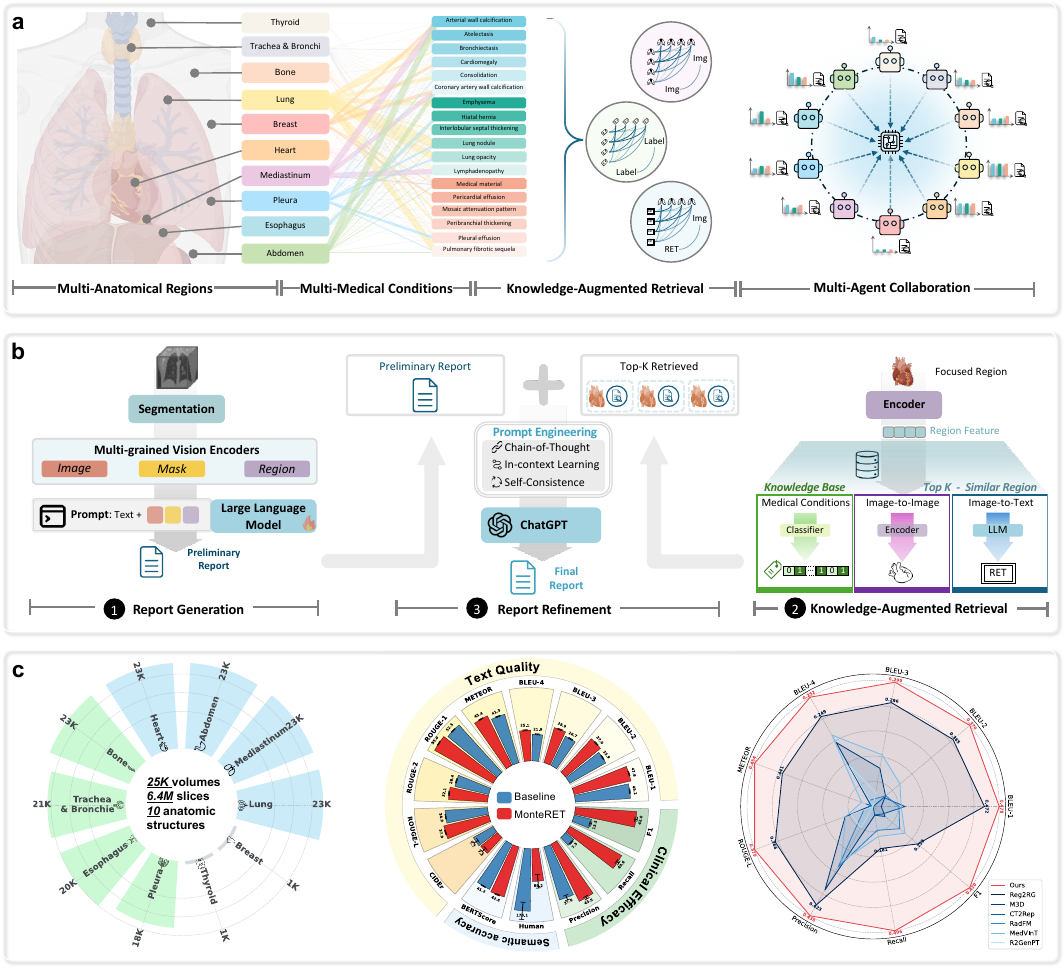}
    \caption{\textbf{Overall study design and model pipeline.} 
    \textbf{a,} Illustration of the multi-granularity modeling of anatomical structures and multi-agent collaboration in report refinement. 
    \textbf{b,} The main components of \ourmodel, including a CT report generation module, a knowledge-augmented retrieval module, and a report refinement module. 
    \textbf{c,} Overview of the main experimental design, including dataset, evaluation metrics, and comparison methods.}
    \label{fig:overview}
\end{figure}

\section{Results}\label{sec:results}
\subsection{Overview of the study}

\ourmodel combines multi-granularity anatomical modeling with knowledge-augmented retrieval (Figure~\ref{fig:overview}a). 
It comprises three components (Figure~\ref{fig:overview}b). 
First, a report generation module leverages a fine-tuned LLM to extract global and region-specific CT features and generate an initial report. 
Second, a knowledge-augmented retrieval module identifies clinically relevant knowledge from prior reports based on predicted medical conditions and region-level visual–textual similarity.
Third, a report refinement module integrates the retrieved knowledge to revise the initial report and enhance the clinical fidelity, completeness, and coherence.

Detailed experimental settings and evaluation metrics are presented in Section~\ref{sec:metrics}.
In short, \ourmodel was trained on a public cohort with 24,128 CT scans from RadGenome-ChestCT~\citep{zhang2024radgenome}, which was built on CT-RATE~\citep{hamamci2024developing} (Figure~\ref{fig:overview}c; Table~\ref{tab:exp_dataset}). 
Each scan was paired with its corresponding radiology report.
RadGenome-ChestCT provides 10 predefined anatomical regions of interest: abdomen, bone, breast, heart, esophagus, lung, mediastinum, pleura, thyroid, and trachea/bronchi. 
Each report sentence is aligned with a relevant anatomical region, resulting in 192,242 region-sentence pairs.

\ourmodel was evaluated on 1,564 held-out CT scans from RadGenome-ChestCT and an external cohort of 82 CT scans from the NewYork-Presbyterian/Weill Cornell Medical Center (NYP/WCM). 
Performance was assessed across three complementary dimensions. 
\emph{Text quality} was measured using standard natural language generation metrics, including BLEU~\citep{papineni2002bleu}, ROUGE~\citep{lin2004rouge}, METEOR~\citep{banerjee2005meteor}, and CIDEr~\citep{vedantam2015cider}. 
\emph{Semantic accuracy} was measured using BERTScore~\citep{Zhang2020BERTScore}, GPT-4o-based evaluation~\citep{openai2024gpt4o}, and blinded expert review by four radiology residents (D.N., H.O., E.T., E.B.) with more than five years of chest imaging experience.
\emph{Clinical efficacy} was evaluated by detecting 18 common medical conditions~\citep{chen2020generating,hamamci2024foundation} in generated and reference reports using a previously trained RadBERT-RoBERTa-4m model~\citep{yan2022radbert}. Results are then reported as micro-averaged precision, recall, and F1-score.
We estimated 95\% confidence intervals (95\% CI) using 1,000 bootstrap replicates and performed paired bootstrap testing for model comparisons.

\subsection{\ourmodel improves report generation}

\begin{figure}[thbp]
\centering
\includegraphics[width=\textwidth]{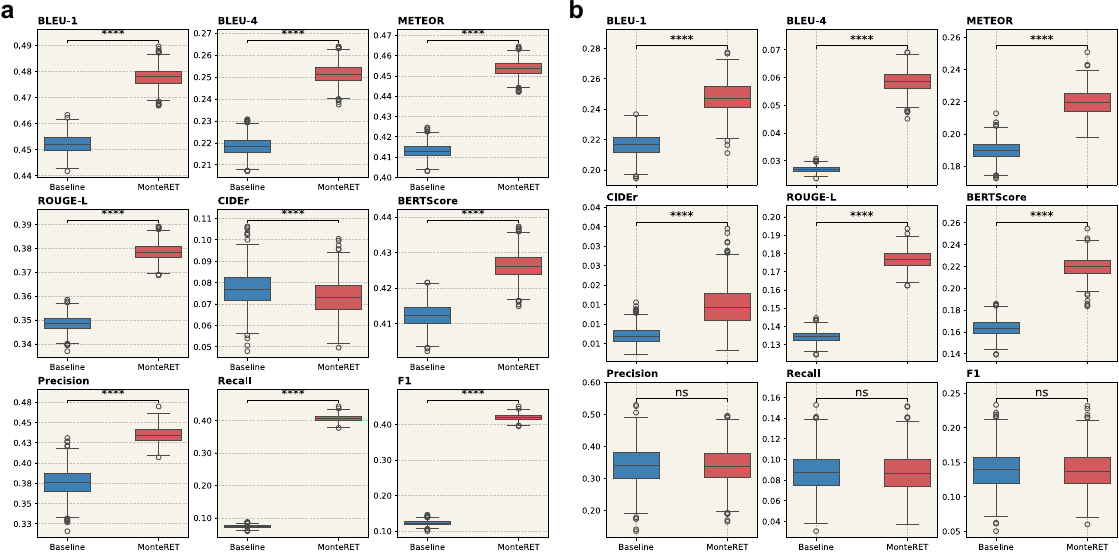}
\caption{\textbf{Comparison of \ourmodel against the baseline model without knowledge-augmented retrieval, vision-language alignment, and region masking.} \textbf{a,} The RadGenome-ChestCT cohort. \textbf{b,} The NYP/WCM cohort. 
Clinical efficacy (precision, recall, F1-score) was reported as the micro-averaged performance across 18 common medical conditions. ns: $p>0.05$, *: $p < 0.05$, **: $p < 0.01$, ***: $p < 0.001$, ****: $p < 0.0001$.}
\label{fig:exp_main}
\end{figure}

We first compared \ourmodel with a matched baseline that used the same pre-trained ViT3D~\citep{wu2025towards} as the visual encoder and LLaMA-2-7B-Chat~\citep{touvron2023llama} as the text decoder, but excluded the report refinement module with knowledge-augmented retrieval, vision-language alignment, and region masking (Figure~\ref{fig:exp_main}; Supplementary Table S1). 

On RadGenome-ChestCT, \ourmodel outperformed the baseline on 8 out of 9 primary metrics (Figure~\ref{fig:exp_main}a). 
For text quality, \ourmodel achieved a BLEU-4 of 0.252, a ROUGE-L of 0.379, and a METEOR of 0.454, corresponding to improvements of 3.3, 2.99, and 4.06 points over the baseline, respectively (all $p<0.0001$; 1 point = 0.01 absolute value).
For semantic accuracy, \ourmodel achieved a BERTScore of 0.426, an improvement of 1.38 points over the baseline ($p<0.0001$).
For clinical efficacy, \ourmodel showed larger gains.
Across the 18 findings, micro-averaged precision, recall, and F1-score reached 0.435, 0.406, and 0.420, respectively.
The largest gain was observed in recall (33.32 points), indicating that retrieval helped recover findings frequently omitted by the baseline.
Notably, the matched baseline already demonstrates competitive performance relative to state-of-the-art approaches, with a BLEU-4 of 0.219 and a ROUGE-L of 0.349.
Against this strong reference, \ourmodel provided additional gains in both language quality and clinical content.

We next evaluated \ourmodel without fine-tuning on the external NYP/WCM cohort (Figure~\ref{fig:exp_main}b).
\ourmodel improves BLEU-1 (3.11 points), BLEU-4 (3.16 points), ROUGE-L (4.25 points), METEOR (3.02 points), CIDEr (0.76 points), and BERTScore (5.64 points) compared with the baseline (all $p<0.0001$).
However, clinical efficacy metrics showed limited improvement in the external cohort, indicating that cross-institution generalization was weaker for clinically oriented endpoints than for text-similarity metrics.
This likely reflects differences in institutional reporting style and the absence of in-domain retrieval examples, highlighting the importance of locally relevant knowledge for retrieval-augmented generation techniques.

\begin{figure}[thbp]
    \centering
    \includegraphics[width=\textwidth]{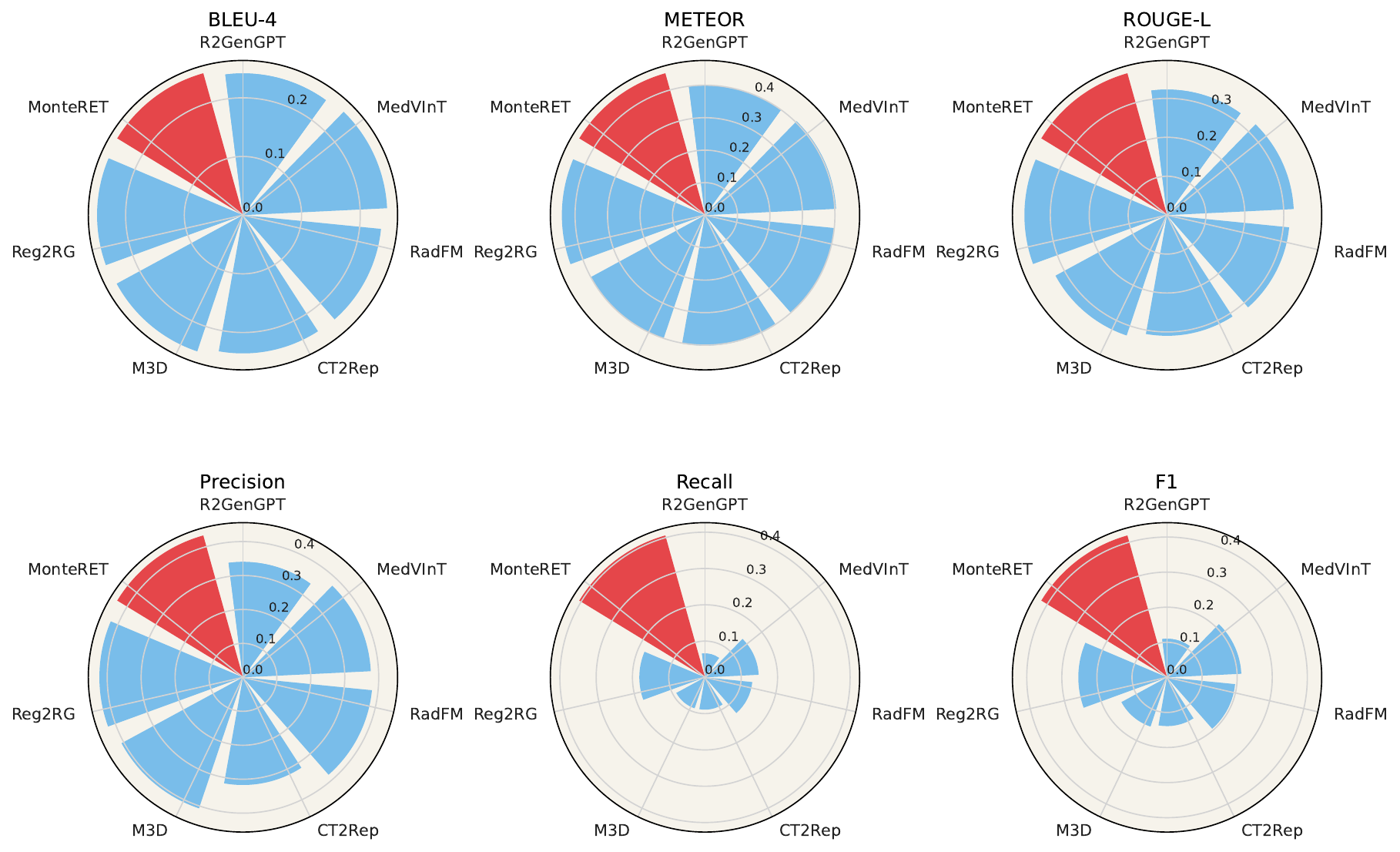}
    \caption{\textbf{Comparison of \ourmodel against several state-of-the-art medical methods on the RadGenome-ChestCT test set.} Compared methods include R2GenGPT~\citep{wang2023r2gengpt}, MedVInT~\citep{zhang2023pmcvqa}, RadFM~\citep{wu2025towards}, CT2Rep~\citep{hamamci2024ct2rep}, M3D~\citep{bai2024m3d}, and Reg2RG~\citep{chen2025large}.}
    \label{fig:exp_sota}
\end{figure}

Finally, we compared \ourmodel with several state-of-the-art methods on the RadGenome-ChestCT test set: R2GenGPT~\citep{wang2023r2gengpt}, MedVInT~\citep{zhang2023pmcvqa}, RadFM~\citep{wu2025towards}, CT2Rep~\citep{hamamci2024ct2rep}, M3D~\citep{bai2024m3d}, and Reg2RG~\citep{chen2025large} (Section~\ref{sec:sota}).
\ourmodel achieved the best overall performance (Figure~\ref{fig:exp_sota}; Supplementary Table S2), with higher BLEU-4 (0.28 points), ROUGE-L (1.20 points), and METEOR (1.29 points) than the strongest competing method Reg2RG.
\ourmodel also improved clinical efficacy, increasing micro-averaged precision (1.17 points), recall (22.51 points), and F1-score (16.68 points).
Together, these results indicate that knowledge-augmented retrieval and multi-grained anatomical region features improve both text quality and clinical completeness.

\FloatBarrier

\subsection{Region-wise evaluation reveals consistent gains with enhanced recall}
\label{sec:region_level_results}

\begin{figure}[!b]
\includegraphics[width=\textwidth]{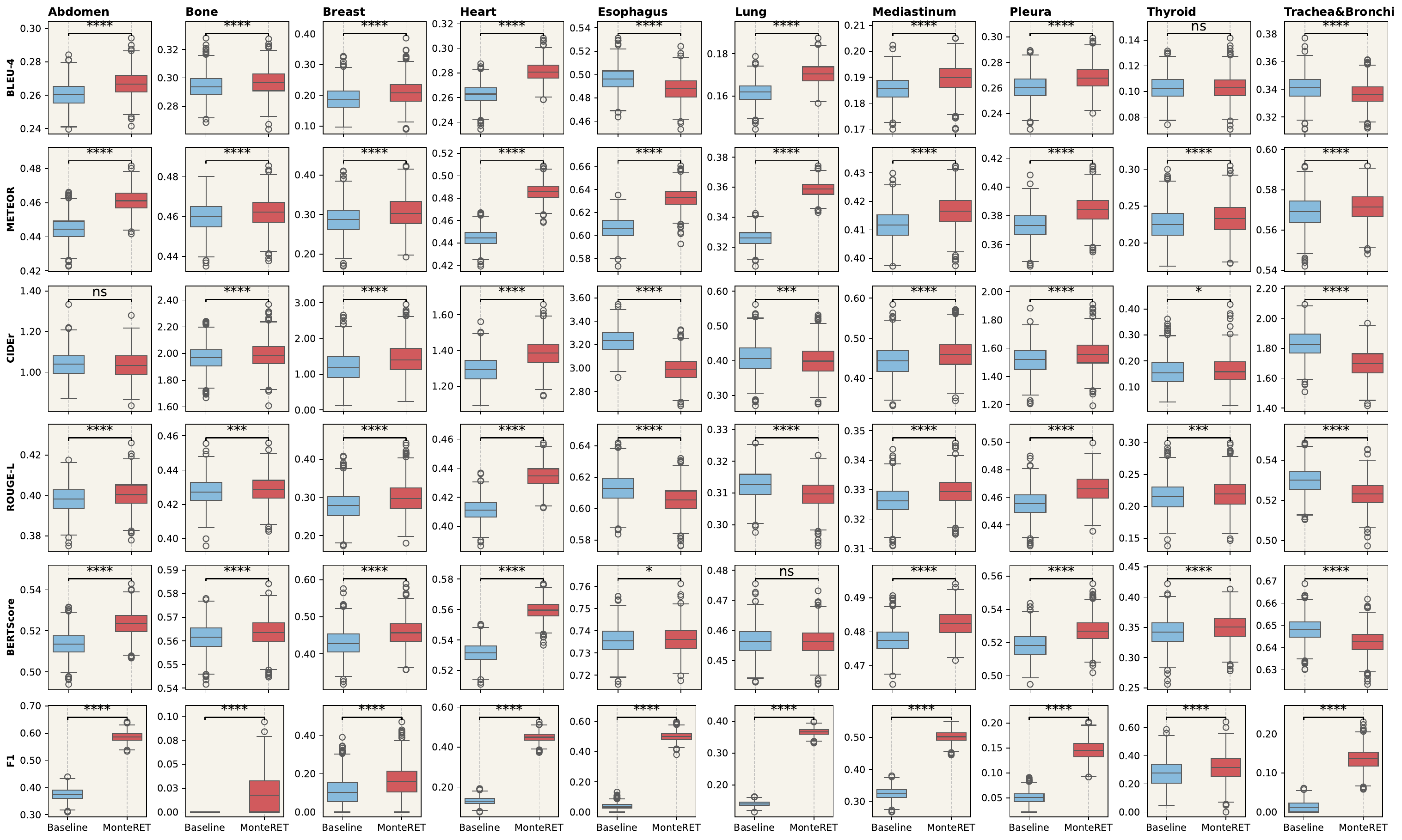}
\caption{\textbf{Comparison of \ourmodel against the baseline method across different anatomical regions.} 
ns: $p>0.05$, *: $p < 0.05$, **: $p < 0.01$, ***: $p < 0.001$, ****: $p < 0.0001$.}
\label{fig:exp_region}
\end{figure}

We next evaluated performance across ten anatomical regions (Figure~\ref{fig:exp_region}; Supplementary Table S3). 
\ourmodel improved most metrics for region-specific description generation. Gains were observed in the abdomen (5/6 metrics), bone (6/6), breast (6/6), heart (6/6), esophagus (3/6), lung (3/6), mediastinum (6/6), pleura (6/6), thyroid (6/6), and trachea/bronchi (2/6).

Across individual metrics, \ourmodel improved BLEU-4 in 8 regions, ROUGE-L in 7 regions, METEOR in all 10 regions, CIDEr in 6 regions, BERTScore in 8 regions, and F1-score in all 10 regions.
The greatest improvements are observed in recall, including gains of 6.53 points in the heart, 21.97 points in the esophagus, and 12.88 points in the lung. These results suggest that the retrieval helped recover clinically relevant findings that were frequently missed by the baseline.

\subsection{Condition-wise evaluation shows broad improvements}
\label{sec:class_level_results}

\begin{figure}[!t]
    \centering
    \includegraphics[width=\textwidth]{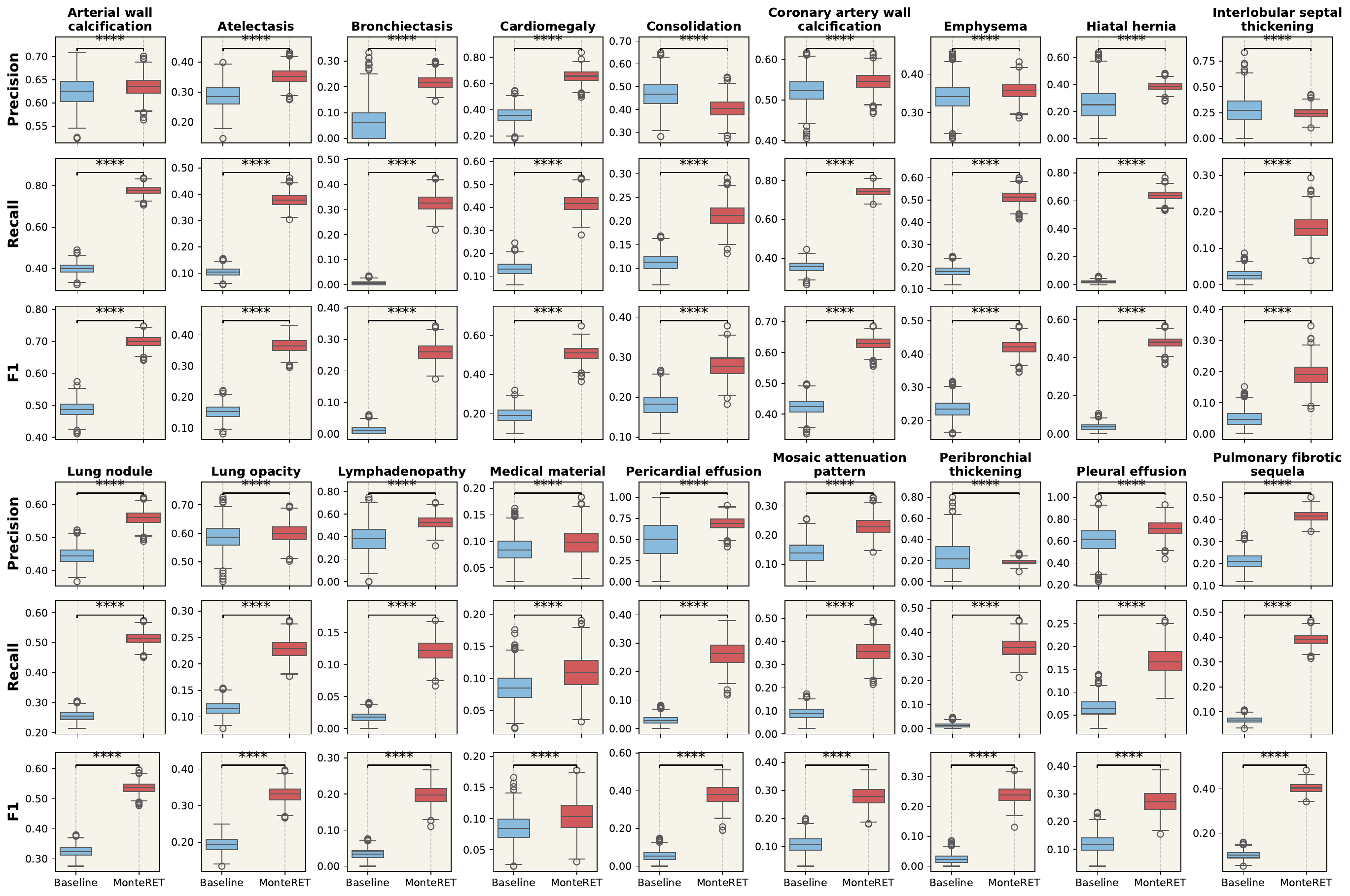}
    \caption{\textbf{Comparison of \ourmodel against the baseline method across different medical conditions and devices.} 
    ns: $p>0.05$, *: $p < 0.05$, **: $p < 0.01$, ***: $p < 0.001$, ****: $p < 0.0001$.}
    \label{fig:exp_class}
\end{figure}

\ourmodel improved recall and F1-score for all 18 conditions and precision for 15 of 18 conditions (Figure~\ref{fig:exp_class}; Supplementary Table S4).
However, performance varied substantially across conditions, with F1-score ranging from 0.104 to 0.700.
Arterial and coronary artery wall calcifications exhibited relatively high precision and recall, likely because these findings are visually distinctive.
In contrast, bronchiectasis showed lower performance, probably due to subtle imaging characteristics and lower prevalence in the training data.

\subsection{\ourmodel partially mitigates class imbalance and region-condition variability}

\begin{figure}[tbhp]
    \includegraphics[width=\textwidth]{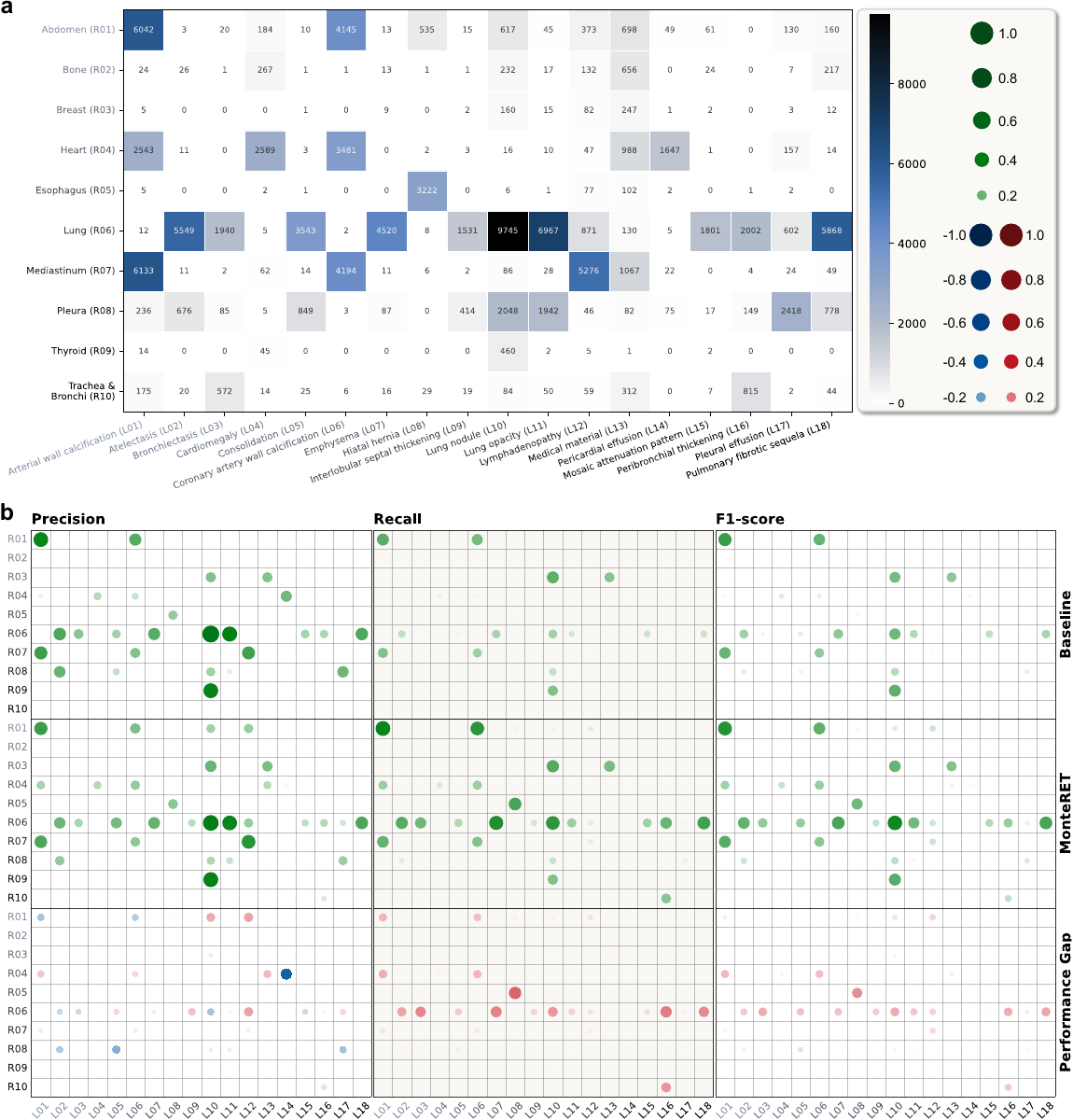}
    \caption{\textbf{Comparison of \ourmodel with the baseline method across 10 anatomical regions and 18 medical conditions.}
    \textbf{a,} Distribution of the samples in the training set of the RadGenome-ChestCT dataset. \textbf{b,} Clinical efficacy of \ourmodel and baseline, measured by micro precision, recall, and F1-score.
    R01: Abdomen, R02: Bone, R03: Breast, R04: Heart, R05: Esophagus, R06: Lung, R07: Mediastinum, R08: Pleura, R09: Thyroid, R10: Trachea \& bronchi.
    L01: Arterial wall calcification, L02: Atelectasis, L03: Bronchiectasis, L04: Cardiomegaly, L05: Consolidation, L06: Coronary artery wall calcification, L07: Emphysema, L08: Hiatal hernia, L09: Interlobular septal thickening, L10: Lung nodule, L11: Lung opacity, L12: Lymphadenopathy, L13: Medical material, L14: Mosaic attenuation pattern, L15: Peribronchial thickening, L16: Pericardial effusion, L17: Pleural effusion, L18: Pulmonary fibrotic sequela.
    }
    \label{fig:exp_cla_region_class}
\end{figure}

The RadGenome-ChestCT training set showed class imbalance across anatomical regions and medical conditions (Figure~\ref{fig:exp_cla_region_class}a). This imbalance makes it challenging for the multi-modal report generation model trained with standard language-modeling loss to identify less prevalent conditions.
\ourmodel was designed to reduce this limitation by combining region-aware modeling with knowledge-augmented retrieval. The condition-guided retrieval module identifies reports containing predicted conditions, increasing the availability of relevant examples for rare or underrepresented findings during inference. The region-level retrieval strategy further constrains this evidence to anatomically matched contexts, helping the model distinguish findings that may otherwise be diluted by dominant lung-related patterns or common normal descriptions.

Consistent with this design, \ourmodel narrowed the gap caused by class imbalance and improved clinical efficacy over the baseline across a broad range of region–condition pairs (Figure~\ref{fig:exp_cla_region_class}b).
Improvements were observed not only in well-represented conditions, but also in less frequent and anatomically challenging settings. The performance-gap panel further shows that \ourmodel reduced many of the baseline's missed findings, with gains most evident in recall and F1-score. For example, in the lung region, \ourmodel identified positive cases of lymphadenopathy, pleural effusion and interlobular septal thickening that were missed by the baseline.

\subsection{Evaluation by human experts and LLM-as-a-Judge}

Four radiology residents (D.N., H.O., E.T., E.B.), each with over 5 years of experience in chest imaging, independently evaluated 50 randomly selected cases. Each case included the reference report and generated reports from \ourmodel and baseline, with the model identity blinded.
Reports were scored on a five-point error severity scale, where 0 indicated ``strongly acceptable''  and 4 ``strongly unacceptable.''
Average Error Severity (AES) was computed by averaging radiology resident ratings across regions~\citep{jeong2024multimodal}. 
Gwet's AC1 coefficients among the radiology residents ranged from 0.656 to 0.952 (Supplementary Figure S1).
Additional details are provided in Supplementary Methods S1.

\begin{figure}[!t]
    \centering
    \includegraphics[width=.8\textwidth]{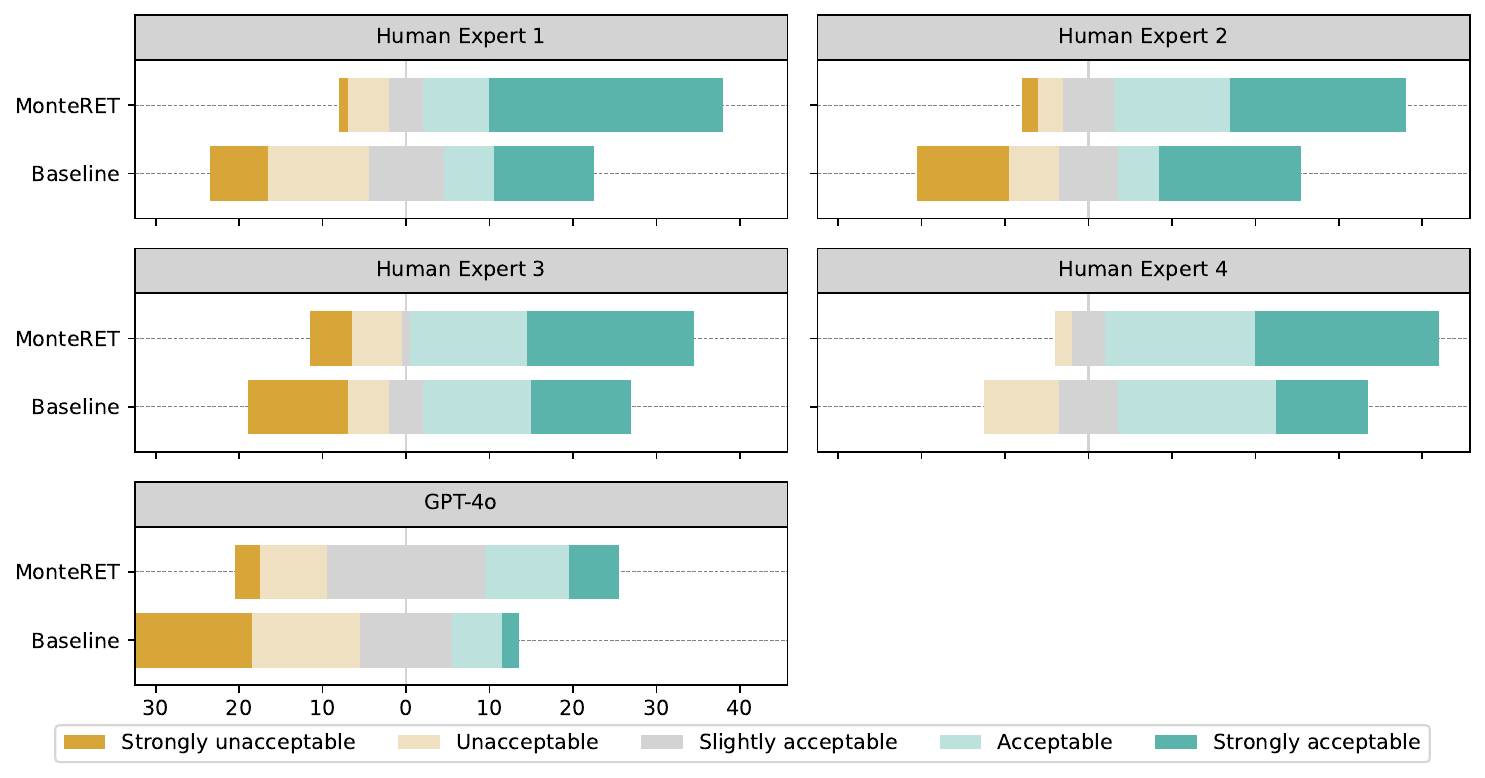}
    \caption{\textbf{Evaluation results from human experts and LLM-as-a-judge (GPT-4o).} Each bar represents the number of cases where the corresponding method was rated as strongly unacceptable, unacceptable, slightly acceptable, acceptable, or strongly acceptable.}
    \label{fig:exp_human}
\end{figure}

\ourmodel achieves lower AES than the baseline for all four reviewers (Figure~\ref{fig:exp_human}).
AES scores for \ourmodel are 0.761, 0.935, 1.174, and 0.696, compared with 1.913, 1.761, 1.826, and 1.304 for the baseline. These differences were statistically significant (all $p < 0.05$).

The same cases were also evaluated using GPT-4o as an LLM-based judge with the same criteria.
GPT-4o assessments were consistent with the radiology residents, with \ourmodel achieving a lower AES than the baseline. However, GPT-4o assigned higher severity ratings overall, potentially suggesting a more conservative evaluation pattern~\citep{Jeblick2024-fe}. 

\subsection{Ablation study}

\subsubsection{Impact of each component in \ourmodel}

\begin{figure}[t]
    \centering
    \begin{minipage}[t]{0.49\textwidth}
        \textbf{a}\par\vspace{2pt}
        \includegraphics[width=\linewidth]{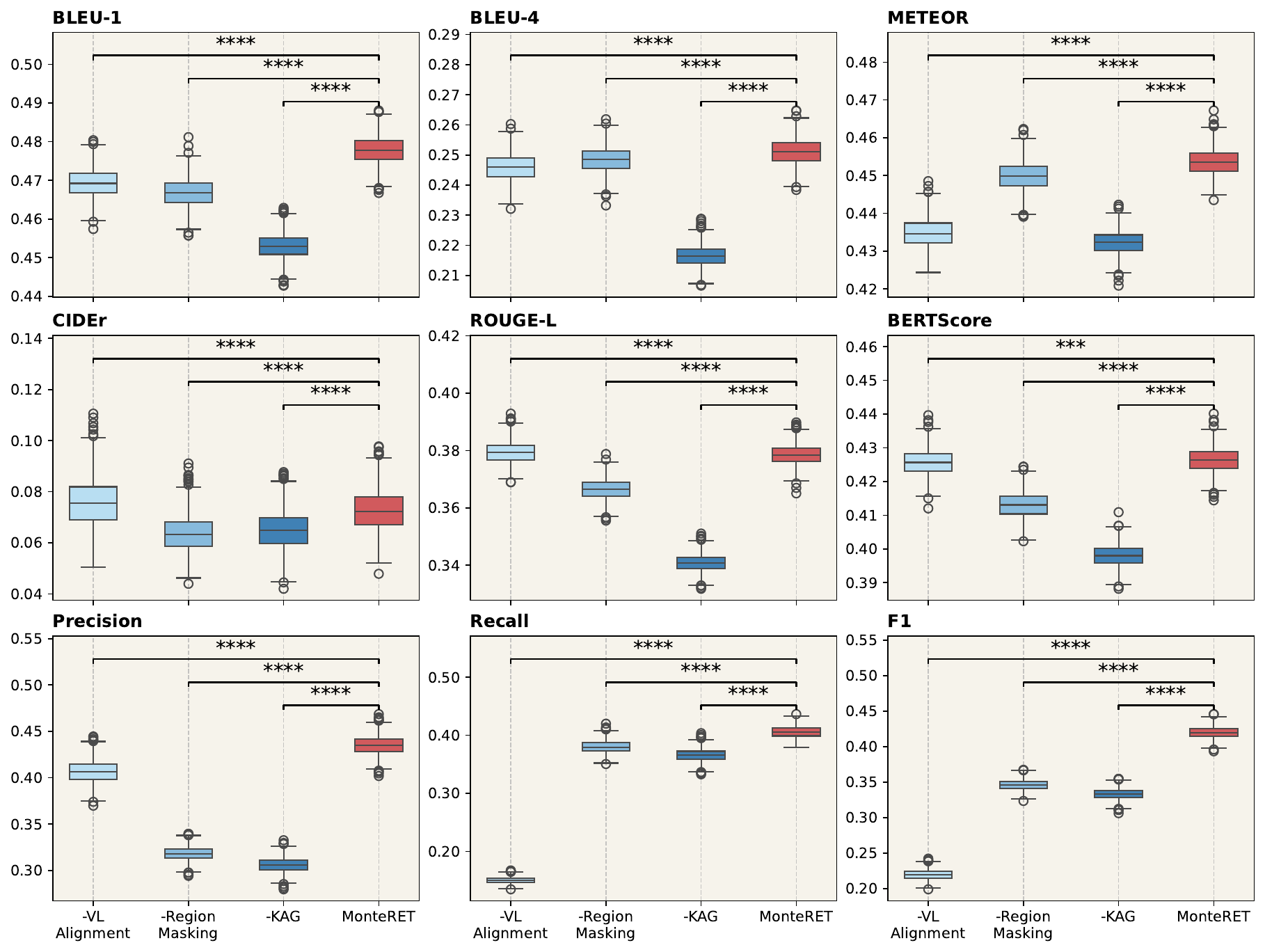}
    \end{minipage}\hfill
    \begin{minipage}[t]{0.49\textwidth}
        \textbf{b}\par\vspace{2pt}
        \includegraphics[width=\linewidth]{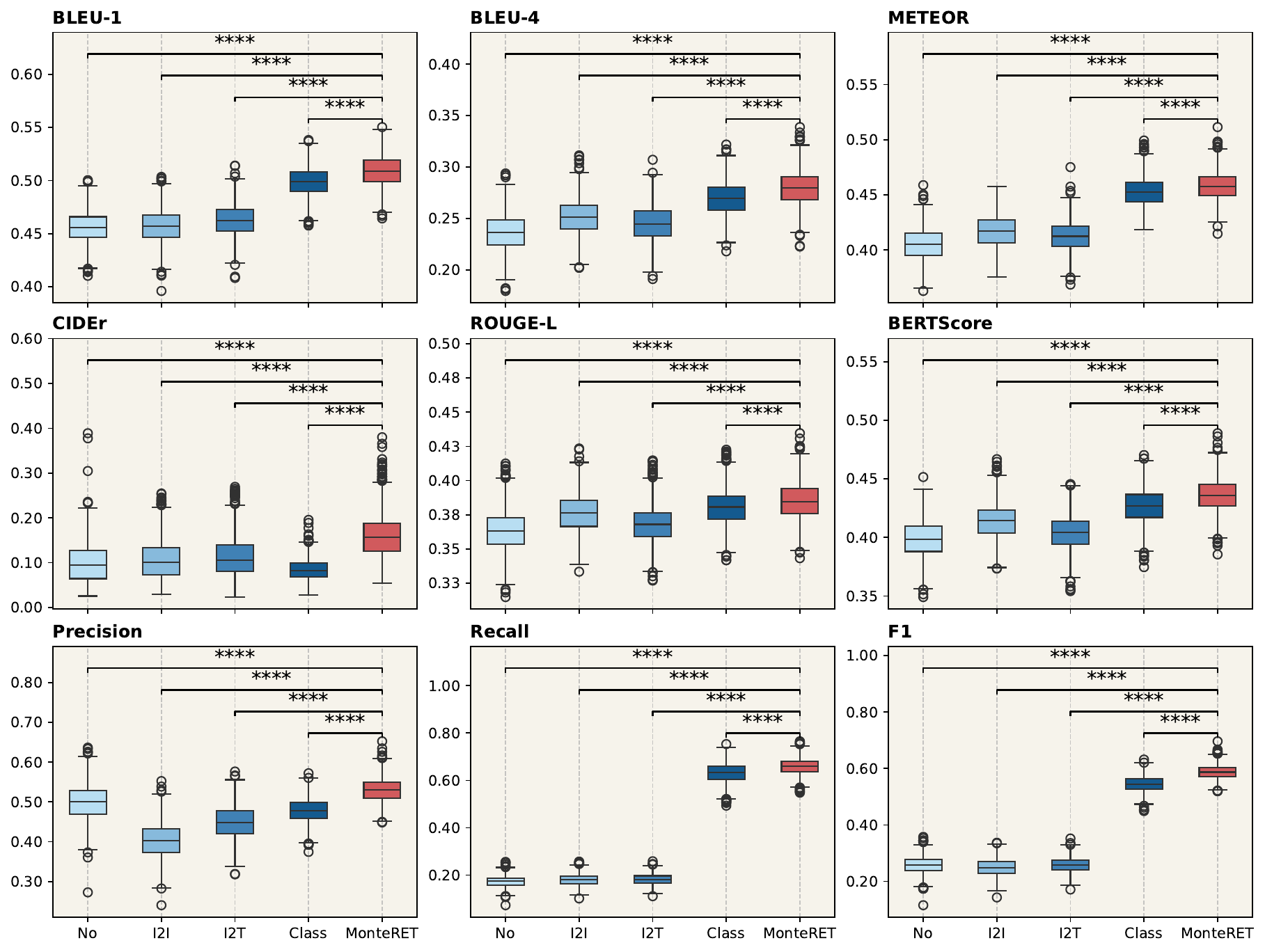}
    \end{minipage}
    \caption{\textbf{Ablation analyses of MonteRET and its knowledge-retrieval strategies.}
    \textbf{a,} Ablation study of the key components. -KAR: removing knowledge-augmented retrieval; -VLA: removing vision-language alignment; -RM: removing region masking.
    \textbf{b,} Comparison of different knowledge retrieval methods. No: without knowledge retrieval; I2I: image-to-image retrieval; I2T: image-to-text retrieval; Class: multi-label classification-based retrieval; MonteRET: proposed knowledge-augmented retrieval module. ns: $p>0.05$, *: $p < 0.05$, **: $p < 0.01$, ***: $p < 0.001$, ****: $p < 0.0001$.}
    \label{fig:exp_ablation}
    \label{fig:exp_retrieve}
\end{figure}

Ablation experiments showed that each component in \ourmodel contributed to performance (Figure~\ref{fig:exp_ablation}a). 
Removing \textbf{knowledge-augmented retrieval} results in the largest decrease across all metrics, especially in recall (25.6 points) and F1-score (20.1 points).
This indicates that external medical knowledge is crucial for identifying clinically relevant findings.
Removing \textbf{vision–language alignment} also degrades both language quality and clinical efficacy, though to a lesser extent. It supports the value of explicit region-level alignment between CT features and text. 
Finally, removing \textbf{region masking} leads to a moderate decrease.
This demonstrates that random region masking enhances the robustness by encouraging the model to use both global and local features.

\subsubsection{Effects of knowledge-augmented retrieval strategies}

We compared four retrieval settings (Figure~\ref{fig:exp_retrieve}b; Supplementary Table S5).

\textbf{No retrieval} (Baseline) generated reports solely from visual CT features. 
\textbf{Image-to-image retrieval} selected anatomically similar chest CT regions from the training set and used their associated reports as context. 
\textbf{Image-to-text retrieval} used a retrieval token from our vision-language alignment module to retrieve semantically relevant CT reports. 
\textbf{Classification-based retrieval} used the predicted conditions to retrieve reports containing similar findings.
\textbf{The full retrieval strategy} combines I2T and class-based retrieval to leverage both semantic alignment and structured clinical condition cues, thereby exploiting the complementary strengths of the two approaches.

Image-to-image and image-to-text retrieval provided limited improvement when used alone, likely because visually similar CT regions do not always share the same clinically relevant findings.
By contrast, classification-based retrieval outperformed both the baseline and other retrieval strategies, indicating that condition-driven retrieval provides clinically meaningful evidence.

\ourmodel achieved the best overall performance, including higher BLEU-4, ROUGE-L, METEOR, CIDEr, precision, recall, and F1-score. 
These findings suggest that, beyond classification-based retrieval, region-level vision–language alignment captures more nuanced details and helps select contextually relevant examples.

\FloatBarrier

\subsubsection{Impact of prompt engineering on report refinement}

\begin{figure}[tb]
    \centering
    \begin{minipage}[t]{0.49\textwidth}
        \textbf{a}\par\vspace{2pt}
        \includegraphics[width=\linewidth]{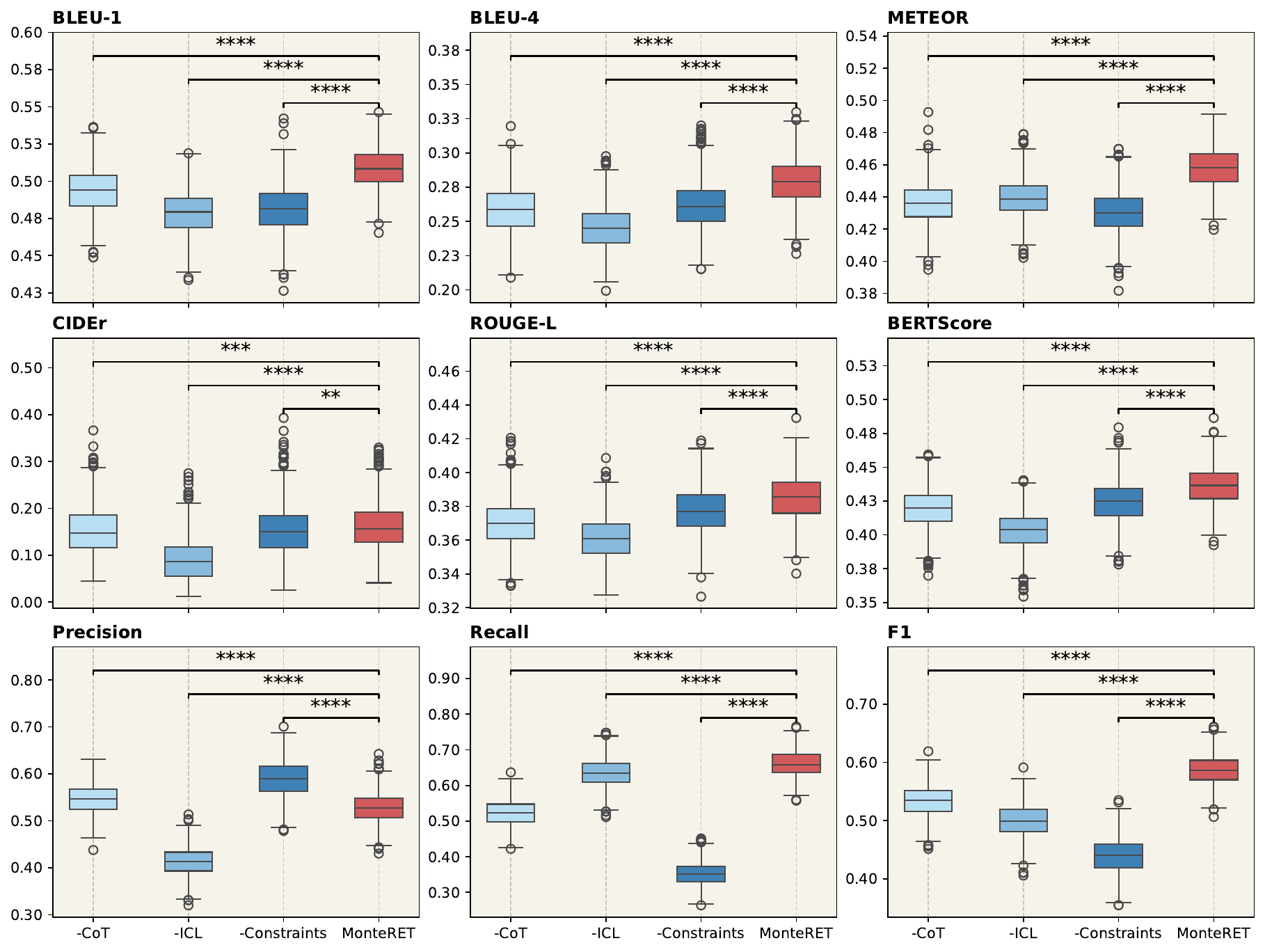}
    \end{minipage}\hfill
    \begin{minipage}[t]{0.49\textwidth}
        \textbf{b}\par\vspace{2pt}
        \includegraphics[width=\linewidth]{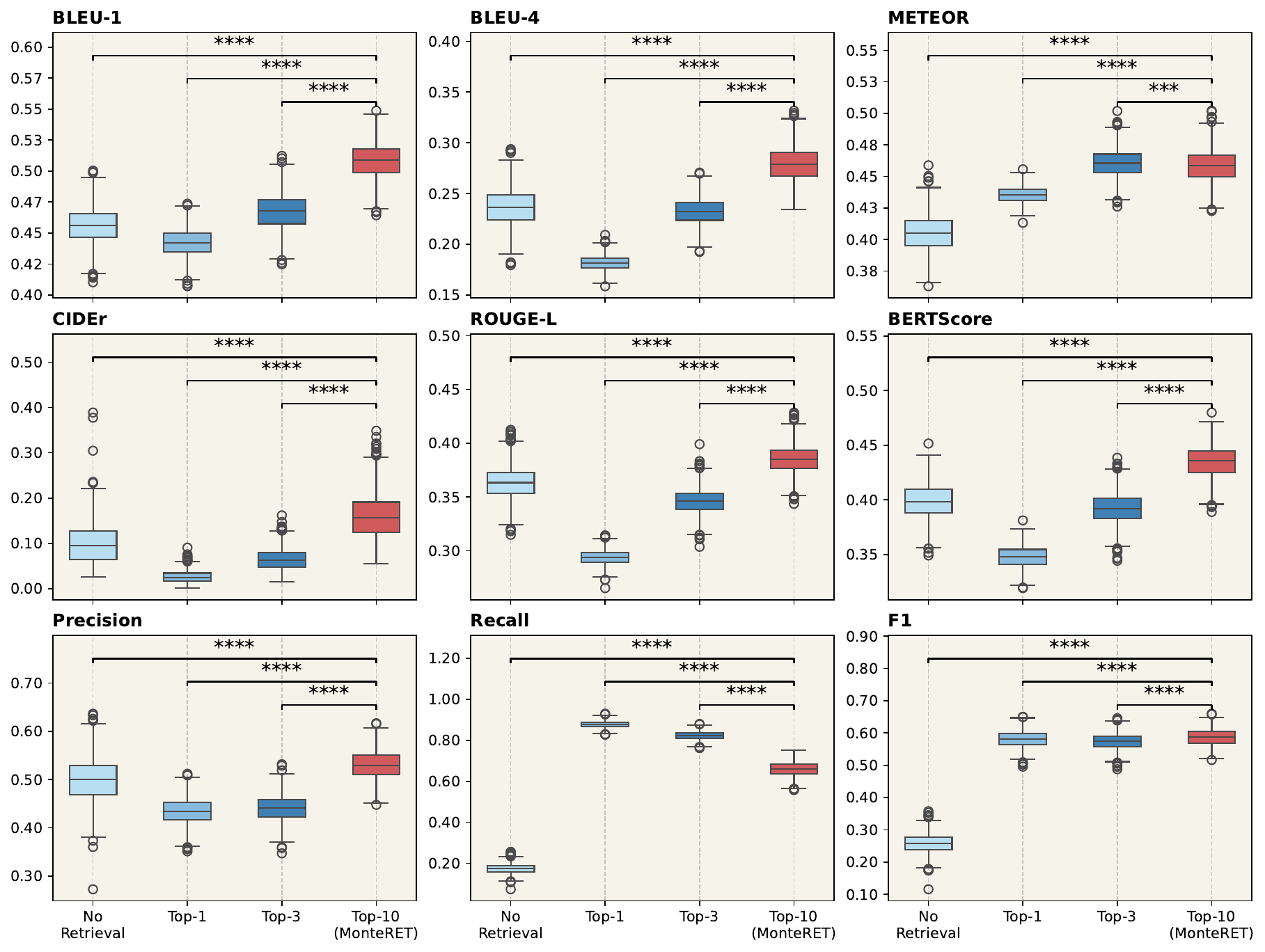}
    \end{minipage}
    \caption{\textbf{Ablation analyses of the report refinement agent.}
    \textbf{a,} Comparison of different prompt engineering strategies. -CoT: removing Chain-of-Thought; -ICL: removing In-Context Learning; -Constraints: removing Medical Domain Knowledge Constraints.
    \textbf{b,} Impact of the number of retrieved candidate reports on report refinement performance.}
    \label{fig:exp_prompt}
    \label{fig:exp_prompt_size}
\end{figure}

We examined the impact of three prompt engineering strategies in the report refinement module (Figure~\ref{fig:exp_prompt}a; Supplementary Table S6).
\textbf{Chain of Thought (CoT)} encourages the LLM to generate intermediate reasoning steps before producing the final report.
\textbf{In-Context Learning (ICL)} provides the LLM with several examples of reference reports to guide the generation process. 
\textbf{Constraints} incorporate specific guidelines in the prompt to ensure that the generated report adheres to clinical standards or formats. For instance, we instruct the model to use the majority vote of the retrieved reports to determine whether each medical condition is present.

Because of LLM usage constraints, experiments were performed on a randomly selected subset of 100 CT scans from the RadGenome-Chest test set.

All three prompt strategies contributed to the performance.
The full one achieved the best overall performance across nearly all metrics. The improvements are especially pronounced in clinical efficacy.
Specifically, removing CoT resulted in a slight performance drop (5.4 points in F1-score, 2.0 points in BLEU-4).
This suggests that while CoT helps structure the report generation process, its impact is relatively modest.
In contrast, removing ICL caused a larger decrease (8.7 points in F1-score, 3.4 points in BLEU-4), indicating that examples can guide report format and style.
Removing the Constraints produced the largest decrease in F1-score (14.7 points), highlighting the importance of domain-specific instructions for maintaining clinical fidelity.

\subsubsection{The impact of the number of retrieved candidate reports}

We further investigated the number of retrieved candidate reports (Figure~\ref{fig:exp_prompt_size}b; Supplementary Table S7).
We varied the size from 1 to 10 and evaluated the performance using the same metrics as before.
Small candidate sets underperformed because they provided insufficient context. A top-10 setting achieved the best balance across language and clinical metrics. Larger retrieval sets may offer richer evidence, but excessive or irrelevant context could reduce model focus.
In contrast, the clinical efficacy metrics increase monotonically with retrieval size.
The precision steadily improved with larger candidate sets, while recall decreased slightly.
This pattern suggests that larger sets enhanced majority voting reliability (higher precision), but may also reduce over-prediction of possible conditions (higher recall but lower NLG metrics).

\FloatBarrier

\subsection{Qualitative analysis}

\begin{figure}[htbp]
    \centering
    \includegraphics[width=.9\textwidth]{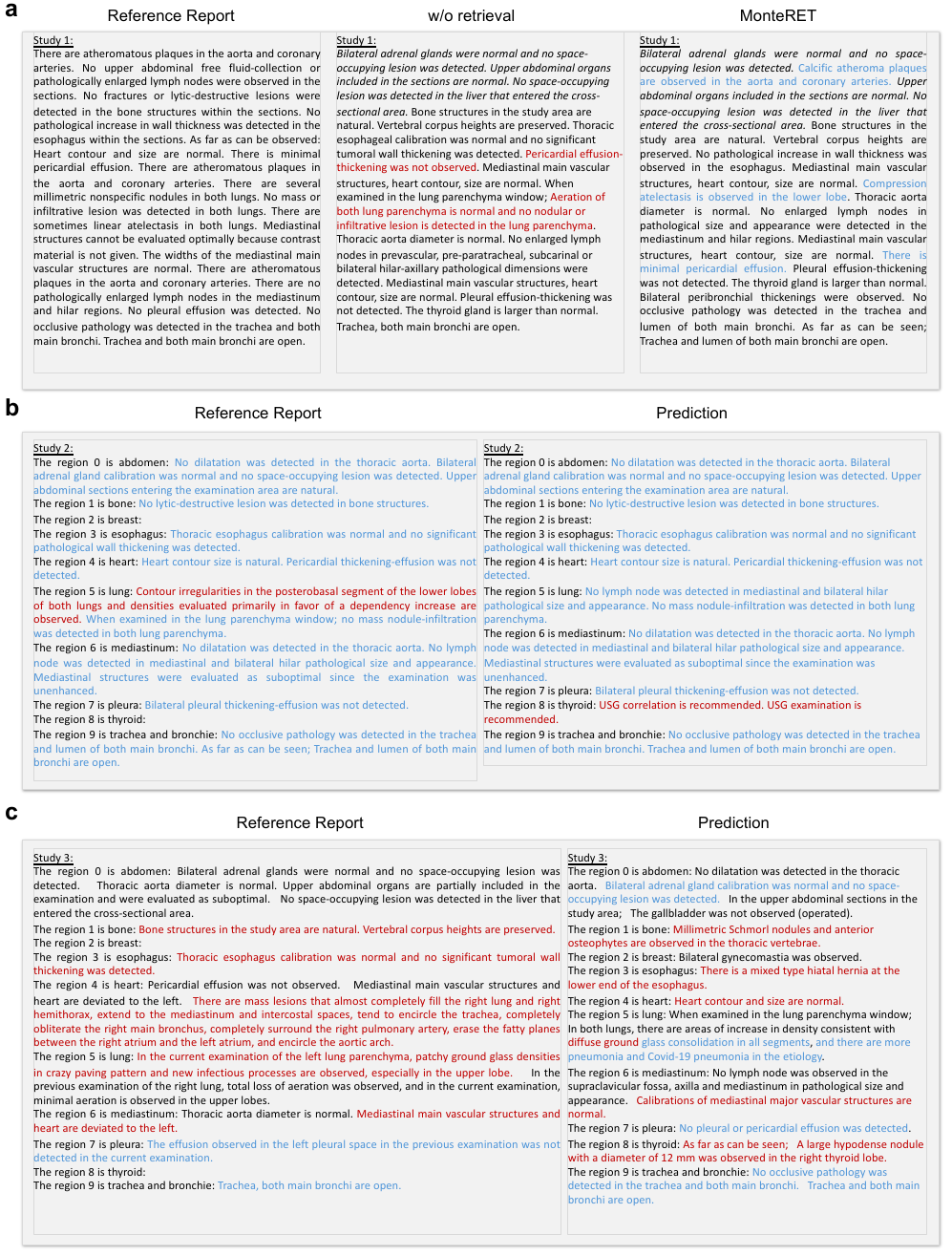}
    \caption{\textbf{Qualitative results of different methods on representative examples.} 
    \textbf{a,} Reference report, and reports generated by the baseline (without the knowledge-augmented retrieval) and \ourmodel. 
    \textbf{b,} Region-level reports from the reference and generated by \ourmodel. 
    \textbf{c,} A failure case. The red and blue colors highlight the errors and correct descriptions, respectively.}
    \label{fig:exp_qualitative}
\end{figure}

Qualitative examples showed that \ourmodel generated more complete and clinically aligned reports than the baseline (Figure~\ref{fig:exp_qualitative}). 
In Figure~\ref{fig:exp_qualitative}a, the baseline (without knowledge-augmented retrieval) missed a minimal pericardial effusion, whereas \ourmodel correctly included this abnormal finding and added relevant vascular calcifications. 
The region-level example in Figure~\ref{fig:exp_qualitative}b shows that \ourmodel generated anatomically structured descriptions consistent with reference reports, including normal heart size and absence of pleural thickening or effusion.

However, errors remained. In one complex case with multiple conditions (Figure~\ref{fig:exp_qualitative}c), \ourmodel generated unsupported diagnoses, such as pneumonia or COVID-19, and incompletely described large mass lesions.
These errors may arise from segmentation challenges in distorted anatomy, variability in the training reports, and next-token prediction bias toward common report patterns.

\FloatBarrier

\section{Discussion}\label{sec:discussion}

Most prior report generation studies have focused on chest X-rays~\citep{Tanno2025-ui,Yang2023-ej,Wang2022-tl,Tanida2023-hi,Liu2021-cr,Li2023-tm,Park2025-zc,Jin2024-sk,yan2022radbert,wang2023r2gengpt}. 
CT report generation remains more challenging because volumetric imaging requires assessment of multiple anatomical structures and subtle regional abnormal findings. Existing methods often generate reports from whole-scan representations, thereby overlooking localized findings. 

In this study, we introduced \ourmodel to address this limitation by explicitly modeling ten anatomical regions and retrieving region-specific knowledge. 

\ourmodel improved report quality and clinical efficacy compared with a matched baseline and several state-of-the-art models (Figure~\ref{fig:exp_main} and Figure~\ref{fig:exp_sota}). Gains were most notable in recall, suggesting that retrieved clinical context helps reduce omitted findings.
This is clinically important because missed findings are a major limitation of automated report generation systems.

We further evaluated \ourmodel at a finer granularity across 10 anatomical regions and 18 medical conditions (Figure~\ref{fig:exp_region}, Figure~\ref{fig:exp_class}, Figure~\ref{fig:exp_cla_region_class}). 
Region- and condition-level analyses showed that \ourmodel attained substantial gains in clinical efficacy across most anatomical regions and conditions. 
Improvements were especially pronounced in diagnostically challenging settings, where subtle imaging cues make accurate report generation more difficult (e.g., Esophagus -- Hiatal hernia and Lung -- Pulmonary fibrotic sequela). 
This advantage became especially evident in class-imbalanced scenarios, where certain medical conditions were underrepresented in the training data. 
In such cases, the knowledge-augmented retrieval module proved critical by using multi-label classification to retrieve clinically relevant examples and region-level vision–language alignment to select anatomically appropriate evidence. 
This additional clinical context improved both the accuracy and specificity of generated reports, particularly for rare or hard-to-diagnose conditions.

Human expert evaluation supported the automated results (Figure~\ref{fig:exp_human}). 
Radiology residents consistently rated \ourmodel reports as more acceptable than baseline reports. 
GPT-4o-based evaluation showed a similar pattern. These findings reinforce the need to supplement automated metrics with expert assessment when evaluating clinical report generation systems.

Ablation studies further showed that the knowledge-augmented retrieval module was the most important contributor to performance (Figure~\ref{fig:exp_ablation}a,b and Figure~\ref{fig:exp_prompt}a,b). Classification-based retrieval helped identify reports containing relevant medical conditions, while vision–language reranking selected examples aligned with the query region. This approach was particularly useful for infrequent findings and imbalanced classes. The refinement module further improved factual consistency by using retrieved evidence under structured clinical constraints.

This study has limitations. 
First, \ourmodel was evaluated only on chest CT and may not generalize directly to other imaging modalities or anatomical domains. 
Second, retrieval performance depends on the quality and relevance of the knowledge database. Rare or emerging conditions may remain difficult to generate accurately if they are underrepresented in the database. 
Third, class imbalance remains a challenge despite the use of condition-guided retrieval. 
Finally, the external validation cohort was relatively small, limiting conclusions about generalizability across institutions.

In summary, \ourmodel demonstrates that anatomical region-aware modeling and knowledge-augmented retrieval can improve chest CT report generation.
The key contribution of MonteRET is the integration of multi-granularity CT representations, condition-guided retrieval and region-specific report refinement within a unified multimodal LLM framework. 
Future work should prospectively evaluate the system, expand external validation, integrate institution-specific retrieval databases, and assess its role in radiologist-AI collaboration.

\newpage

\section{Methods}\label{sec:methods}

\subsection{Ethical approval}

Data collection protocols and the use of CT scans, radiological reports, and clinical information were approved by the Institutional Review Board (IRB) of Weill Cornell Medicine (IRB Protocol\#: 24-02027110). The IRB granted a waiver of informed consent for the retrospective use of CT scans and associated clinical data. All procedures were conducted in accordance with institutional ethical standards and adhered to the principles outlined in the Declaration of Helsinki.

\begin{table}[thbp]
    \centering
    \caption{Characteristics of the RadGenome-Chest cohort and the NYP/WCM cohort.}
    \label{tab:exp_dataset}
    \begin{tabular}{lrrr}
    \toprule
 & \multicolumn{2}{c}{RadGenome-Chest CT scans, No.} & NYP/WCM CT scans, No.\\
\cmidrule(r){2-3}\cmidrule{4-4}
Characteristic &  Train & Test & Test\\
& (n=24,128) & (n=1,564) & (n=82)\\
\midrule
Participants, No. & 20,000 & 1,304 & 82 \\
\arrayrulecolor{gray}\hline
Age, mean (SD), y & 48.74 (17.28) & 48.39 (16.87) & 69.32 (19.17) \\
\hline
Sex \\
~~ Male & 14,099 & 910 & 51\\
~~ Female & 10,029 & 654 & 31\\
\hline
Regions & \\
~~Abdomen & 23,553 & 1,518 & 69\\
~~Bone & 23,479 & 1,509 & 70\\
~~Breast & 1,080 & 58 & --\\
~~Heart & 23,289 & 1,433 & 78\\
~~Esophagus & 20,693 & 1,328 & 11\\
~~Lung & 23,741 & 1,514 & 66\\
~~Mediastinum & 23,684 & 1,523 & 52\\
~~Pleura & 18,156 & 1,172 & 3 \\ 
~~Thyroid & 1,093 & 51 & 3\\
~~Trachea/bronchi & 21,951 & 1,417 & 26\\
\hline
Medical conditions\\
~~Arterial wall calcification & 6,607 & 423 & 38 \\
~~Atelectasis & 6,005 & 359 & 36 \\
~~Bronchiectasis & 2,341 & 163 & 15 \\
~~Cardiomegaly & 2,533 & 159 & 5 \\
~~Consolidation & 4,066 & 280 & 28 \\
~~Coronary artery wall calcification & 5,747 & 348 & 40 \\
~~Emphysema & 4,558 & 304 & 16 \\
~~Hiatal hernia & 3,391 & 197 & 10 \\
~~Interlobular septal thickening & 1,887 & 119 & 14 \\
~~Lung nodule & 10,999 & 697 & 29 \\
~~Lung opacity & 8,944 & 607 & 41 \\
~~Lymphadenopathy & 5,839 & 343 & 21 \\
~~Medical material & 2,846 & 149 & 25 \\
~~Mosaic attenuation pattern & 1,788 & 124 & 7 \\
~~Peribronchial thickening & 2,566 & 178 & 6 \\
~~Pericardial effusion & 1,641 & 106 & 14 \\
~~Pleural effusion & 2,628 & 179 & 10 \\
~~Pulmonary fibrotic sequela & 6,175 & 399 & 10 \\
\bottomrule
    \end{tabular}
\end{table}

\subsection{Study design and study population}

For the RadGenome-ChestCT dataset, we follow the same data split as in~\citep{chen2025large}, which includes 24,128 and 1,564 CT scans for training and testing, respectively.
Demographic information for patients in each subset is summarized in Table~\ref{tab:exp_dataset}.
The average age of patients is 48.74 years (standard deviation: 17.28 years), with a sex distribution of 42\% female and 58\% male.
Detailed inclusion and exclusion criteria for the RadGenome-ChestCT dataset can be found in~\citep{hamamci2024developing}.

For external testing, we initially identified 99 CT examinations from 99 patients performed at NewYork-Presbyterian/Weill Cornell Medical Center in 2021. 
We then restricted the cohort to non-contrast chest CT studies with corresponding radiology reports. Examinations were excluded if they represented other body regions, contrast-enhanced chest CT, combined multi-region studies, or image files that were not compatible with the preprocessing pipeline. This yielded 82 examinations for the final analysis.
The average age of patients in this dataset is 69.32 years (standard deviation: 19.17 years), comprising 51 females (62\%) and 31 males (38\%).
The demographic information of patients in this dataset is summarized in Table~\ref{tab:exp_dataset}.
The radiological reports contain several sections, including \textit{Exam}, \textit{History},\textit{Technique}, \textit{Comparison}, and \textit{Findings}.
In this study, we focus on generating the \textit{Findings} section, which provides a detailed description of the observed anatomical structures and any abnormalities.
We preprocessed the reports by removing personally identifiable information, then standardized only the report format using GPT-4.1. Specifically, GPT-4.1 was used to convert heterogeneous report layouts into a consistent plain-text \textit{Findings} section by normalizing section headers, punctuation, spacing, and line breaks, while removing non-diagnostic template artifacts.\footnote{The GPT-4.1 API was used with Azure OpenAI Service, which maintains strict data security and privacy standards.} 
The revised reports were manually reviewed to ensure that the original clinical content and diagnostic information were preserved. 

We preprocessed the CT scans by resampling them to a uniform voxel spacing of 1$\times$1$\times$3 mm and resizing them to a fixed dimension of 256$\times$ 256$\times$64 voxels, centered on the lung region.
We applied intensity normalization and clipping to a range of [-1000, 1000] Hounsfield Units (HU) to enhance the contrast of lung parenchyma and relevant anatomical structures.

\FloatBarrier

\subsection{The development of \ourmodel}
\begin{figure}[thbp]
    \centering
    \includegraphics[width=\textwidth]{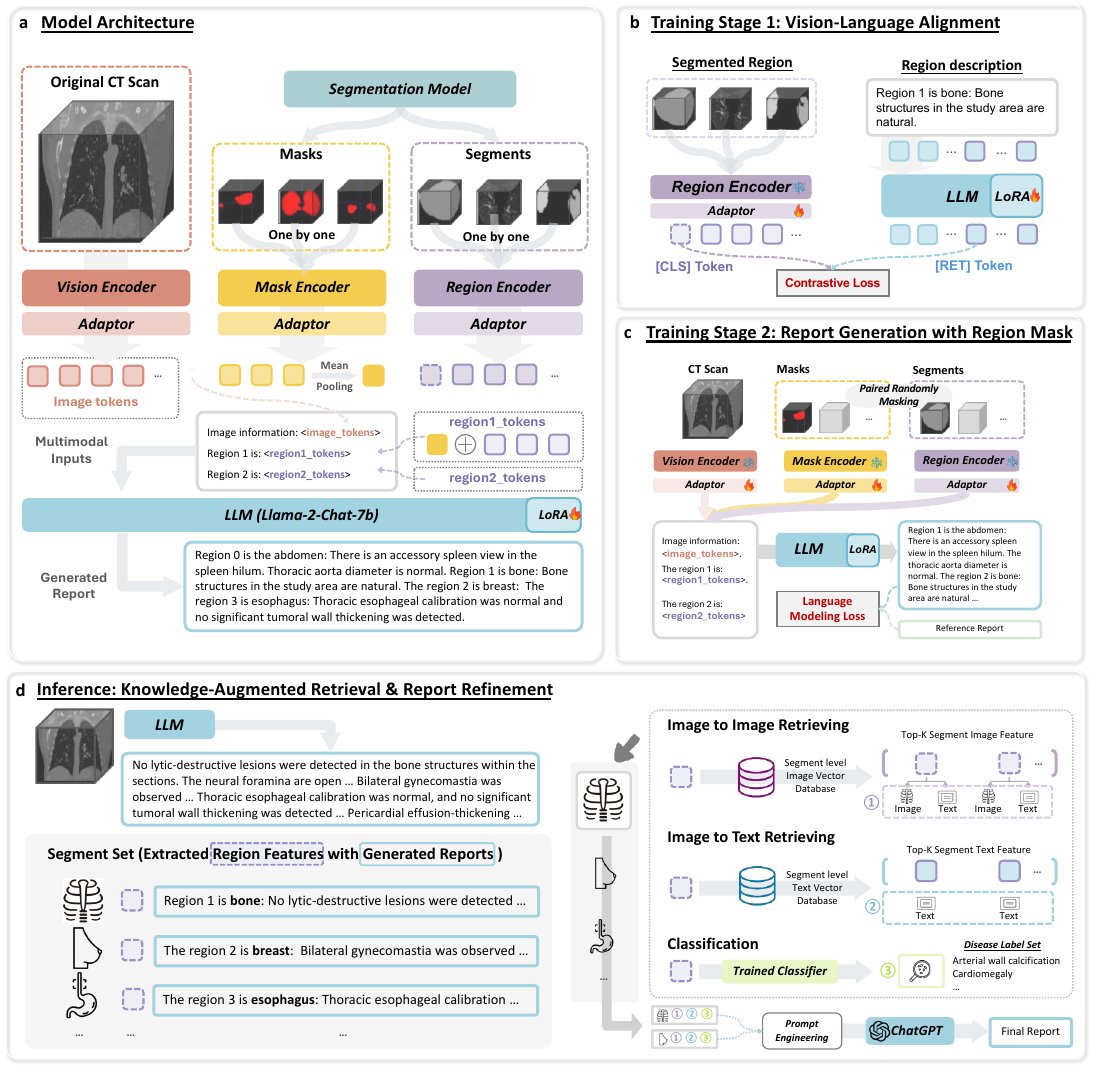}
    \caption{Detailed illustration of \ourmodel. \textbf{a,} The architecture of \ourmodel, which consists of a multi-grained feature extractor and a fine-tuned large language model. \textbf{b,} The first stage of model training, where region-level vision and language features are aligned using contrastive learning. \textbf{c,} The second stage of model training, where the large language model is fine-tuned to generate radiological reports based on the extracted vision features and additional instruction prompts. \textbf{d,} During inference, the generated report is further refined by a multi-agent system using a knowledge-augmented retrieval mechanism.}
    \label{fig:method2}
\end{figure}

In this study, we introduce \ourmodel, a framework that leverages retrieved knowledge to enhance the capabilities of multimodal LLMs in chest CT report generation. 
\ourmodel is composed of three main components: (1) a multimodal network that uses multi-grained visual features for radiological report generation (Figure~\ref{fig:method2}a-c), (2) a knowledge-augmented retrieval module that systematically retrieves pertinent CT reports informed by both high-level medical conditions and fine-grained anatomical visual characteristics (Figure~\ref{fig:method2}d-e), and (3) a report refinement agent that incorporates the retrieved knowledge to iteratively enhance the quality of the generated reports. 
The training configuration and hyperparameters for each component are detailed in Supplementary Methods S2.

\subsubsection{Report generation agent with multi-grained features}
\paragraph{Multimodal network with multi-grained visual features}
The multimodal network within \ourmodel is designed to efficiently process and integrate visual information from CT scans, as shown in Figure~\ref{fig:method2}a. 
It extracts multi-grained features that encompass both global representations and localized anatomical details. 
Global features encapsulate the overall context of the CT scan, while local features are tuned to salient anatomical structures or regions of interest.
These complementary visual representations are subsequently integrated and conveyed to an LLM for generating radiological reports. 
Formally, given a medical image $\mathcal{I}$, the objective is to generate a sequence of $N$ text tokens $\mathcal{T} = \{t_1, t_2, \ldots, t_N\}$ conditioned on the visual characteristics of the input image, defined as

\begin{equation}
    p(\mathcal{T}|\mathcal{I}) = \prod_{i=1}^{N} p(t_i|t_{<i}, \mathcal{I}).
\end{equation}

\ourmodel ($\Phi_{\text{\ourmodel}}$) is then trained to minimize the negative log-likelihood of the correct next token given the previous ones and the image with a regression loss

\begin{equation}
    \mathcal{L}_{\text{reg}} = -\sum_{i=1}^{N} \log \Phi_{\text{\ourmodel}}(t_i|t_{<i}, \mathcal{I}).
\end{equation}

Our multi-grained visual features comprise three components: the global image feature, the anatomical region feature, and the anatomical mask feature. 
These complementary sources collectively furnish a holistic representation of the medical image. 
Specifically, the global image feature captures the broader contextual landscape within the CT scan, the anatomical region feature focuses on delineated regions of interest (ROIs), and the anatomical mask feature delineates the precise spatial location and morphological characteristics of anatomical structures.
To delineate anatomical regions ($\mathcal{R}$) and generate corresponding masks ($\mathcal{M}$), we leverage the SAT model~\citep{zhao2025large-vocabulary}, which enables accurate segmentation of anatomical entities within CT scans.

For the image and region encoders ($\Phi_{I}$, $\Phi_{R}$), we employ the ViT3D architecture, pre-trained on RadFM~\citep{wu2025towards}, owing to its strong ability to capture the complex 3D spatial relationships inherent in volumetric medical images. 
The parameters of both encoders are kept frozen during training to preserve the integrity of the pre-trained features. 
A lightweight adapter then projects the concatenated visual feature vectors into a lower-dimensional embedding space, ensuring compatibility with the subsequent language model.
For the mask encoder ($\Phi_{M}$), we use a lightweight Vision Transformer (ViT) to encode sparse, localized spatial information from anatomical masks.

\paragraph{Region-wise vision-language alignment}
To enhance the model's ability to associate distinct anatomical regions with their corresponding textual descriptions, we introduce a region-wise vision-language alignment (VLA) strategy within the training paradigm (Figure~\ref{fig:method2}b). 
Specifically, we augment the LLM's vocabulary with a novel retrieval token, \texttt{[RET]}, serving as an explicit marker for each anatomical region.
This retrieval token establishes a direct, interpretable link between the visual features extracted from a given region and its corresponding segment in the textual report. 
By doing so, the model is empowered to selectively attend to the most pertinent anatomical context during report generation and retrieve semantically similar reports from the database if required.
Departing from conventional approaches that utilize a single token to represent the entirety of a CT scan~\citep{Jin2024-sk}, our method employs distinct \texttt{[RET]} tokens for each anatomical region (e.g., \texttt{[RET\textsubscript{1}]}, \texttt{[RET\textsubscript{2}]}, ..., \texttt{[RET\textsubscript{n}]}), thereby enabling a fine-grained alignment between visual and textual modalities at the region level.

Given a region-level image-text pair $(\mathcal{R}_i, \mathcal{T}_i)$ corresponding to anatomical region $i \in \{1, 2, \ldots, n\}$ $(n = 10)$ (we omit the subscript $i$ for simplicity), we first derive the visual representation $\Phi_{R}(\mathcal{R})$ via the region encoder, alongside the corresponding textual feature $\Phi_{T}(\mathcal{T})$ using the text decoder.
Then, both representations are mapped into a unified embedding space through modality-specific linear projection heads, denoted as $f_R$ and $f_T$ for the visual and textual modalities, respectively. 
The semantic alignment between region and text features is then assessed by computing their cosine similarity
\begin{equation}
    \text{sim}(\mathcal{R}, \mathcal{T}) = 1 - \frac{f_R(\Phi_{R}(\mathcal{R})) \cdot f_T(\Phi_{T}(\mathcal{T}))}{\|f_R(\Phi_{R}(\mathcal{R}))\| \|f_T(\Phi_{T}(\mathcal{T}))\|},
\end{equation}
where $\|\cdot\|$ denotes the $L_2$ norm.
To align region-level visual and textual features, we minimize an InfoNCE-style loss~\citep{Sohn2016-mc} over both text-to-image and image-to-text retrieval
\begin{equation}
    \mathcal{L}_{\text{align}} = -\frac{1}{2N} \left( \sum_{i=1}^{N} \log \frac{\exp(\text{sim}(\mathcal{R}_i, \mathcal{T}_i)/\tau)}{\sum_{j=1}^{N} \exp(\text{sim}(\mathcal{R}_i, \mathcal{T}_j)/\tau)} + \sum_{i=1}^{N} \log \frac{\exp(\text{sim}(\mathcal{T}_i, \mathcal{R}_i)/\tau)}{\sum_{j=1}^{N} \exp(\text{sim}(\mathcal{T}_i, \mathcal{R}_j)/\tau)} \right),
\end{equation}
where $\tau$ is a learnable temperature that controls the sharpness of the similarity distribution, and
$N$ is the number of region-report pairs in a training batch.

\paragraph{Region mask for balanced contextual learning}
In clinical settings, radiologists routinely rely on both the global context of the entire image and the fine-grained details of specific anatomical structures to inform their diagnoses. 
However, indiscriminately supplying the model with all available anatomical masks risks an undue emphasis on localized features, potentially at the expense of global contextual comprehension.
To address this challenge, we propose a \textbf{region mask} strategy during training (Figure~\ref{fig:method2}c). 
Analogous to the masked language modeling approach employed in BERT~\citep{devlin2019bert}, we randomly mask a subset of anatomical region features with a predetermined probability (empirically set to 0.1) in each training iteration. 
This encourages the model to jointly leverage both global and local cues for report generation, thereby enhancing its capacity to contextualize anatomical findings within the broader image landscape.
Moreover, this strategy increases the model's robustness, enabling it to effectively handle missing or incomplete anatomical information that may arise from segmentation limitations.

\paragraph{Report generation with LLM}
Upon extracting multi-grained visual features, they are integrated to yield a unified and comprehensive representation of the CT scan. 
This fused representation subsequently informs the construction of a text prompt ($\mathcal{P}$), which is fed to the LLM for report generation.
To incorporate multi-grained visual features alongside the textual tokens, we project the visual encodings into specialized placeholder tokens (e.g., \texttt{[image]} and \texttt{[region]}) embedded within a prompt template, $\mathcal{P}_T$. 
The integration process is facilitated by learnable embedding matrices, $W_I$ and $W_R$, which transform the visual features into corresponding token embeddings (see Prompt~\ref{prompt:template t}). 

\begin{prompt}[An example of a prompt template $\mathcal{P}_T$]
\label{prompt:template t}
The global information is provided as the context: \texttt{[image]}. \\
The region 1 is \texttt{[region-1]}. \\
The region 2 is \texttt{[region-2]} \\
... \\
The region 10 is \texttt{[region-10]}. \\
Given the provided global and regional information from this CT scan, please generate a comprehensive medical report for each region.
\end{prompt}

The final prompt, $\mathcal{P}$, is constructed as follows:
\begin{align}
    \mathcal{P} &= [\mathcal{P}_T; \mathcal{P}_I; \mathcal{P}_R],\\
    \mathcal{P}_I &= W_I \Phi_{I}(\mathcal{I}),\\
    \mathcal{P}_R &= W_R (\Phi_{R}(\mathcal{R}) \oplus \Phi_{M}(\mathcal{M})).
\end{align}
Here, $\mathcal{P}_I$ and $\mathcal{P}_R$ denote specialized tokens that encapsulate global image features and anatomical region-specific representations, respectively. 
For each region, the corresponding visual feature is concatenated ($\oplus$) with its associated mask feature, thereby capturing a comprehensive depiction of the underlying anatomical structures, including shape, size, and detailed contextual information.
We employ LLaMA-2-7B-Chat~\citep{touvron2023llama} as the backbone language model. 
To facilitate efficient adaptation to the report generation task without necessitating full model parameter updates, we leverage Low-Rank Adaptation (LoRA)~\citep{hu2022lora} to fine-tune the LLaMA-2 architecture.
The LoRA configuration is set to a rank of 8 and a scaling factor of 32, empirically balancing the trade-off between model expressiveness and computational efficiency.

\subsubsection{Knowledge-augmented retrieval via retrieval and reranking}
To improve the completeness and consistency of the generated reports, we introduce a knowledge-augmented retrieval mechanism that extracts relevant CT reports from a curated repository (Figure~\ref{fig:method2}d). 
This retrieval pipeline leverages not only high-level medical diagnoses but also fine-grained anatomical visual features, thereby ensuring that the incorporated knowledge is both contextually pertinent and clinically informative.
The retrieval procedure is conducted independently for each anatomical region, resulting in a cohort of 10 specialized retrieval agents, each devoted to a distinct anatomical area. 
Specifically, for each region, we employ a two-stage retrieval strategy consisting of initial retrieval and subsequent re-ranking.
In the first stage, a medical condition classifier is realized by fine-tuning the region encoder ($\Phi_{R}$) with linear probe layers to predict the presence of 18 prevalent medical conditions within anatomical regions on chest CT images. 
The classifier is optimized using a class-balanced cross-entropy loss that adaptively weights each medical condition based on its frequency in the training cohort. 
During inference, the classifier predicts the presence of 18 medical conditions for each anatomical region, and these predictions serve as queries to retrieve existing reports from a database documenting the same conditions.
Formally, the retrieval process for a query anatomical region is defined as:
\begin{equation}
    S_{\text{ret}}^i = \min_K \|C(q) - C(i)\|,
\end{equation}
where $C(\cdot)$ denotes the medical condition classifier, $q$ is the query region, and $i$ represents each report in the database.
$S_{\text{ret}}^i$ is the similarity score between the query region and the $i$-th report in the database.
The top-$K$ reports with the highest similarity scores are selected as the initial retrieval results.
In the second stage, a region-wise vision-language alignment (VLA) model is deployed to re-rank the initially retrieved reports according to their semantic congruence with the anatomical region. 
Retrieval tokens (e.g., \texttt{[RET]}) embedded within the prompt function as queries to identify and prioritize reports that are semantically aligned with the anatomical context. 
The similarity scores computed by the VLA model facilitate the reordering of retrieved reports, ensuring that the most relevant and contextually appropriate exemplars are preferentially selected for subsequent integration into the report refinement process.

\subsubsection{Report refinement agent with prompt engineering}
In \ourmodel, we implement a rewriting agent that enhances report generation by judiciously leveraging multi-granularity information from a curated repository of CT reports. 
This module iteratively refines the initial output by incorporating relevant domain knowledge from the database, thereby elevating the quality and completeness of the generated report.
Concretely, given an initial report $\mathcal{T}_{\text{init}}$ produced by the multimodal network, the rewriting agent synthesizes the final report $\mathcal{T}_{\text{final}}$ by integrating the top-$K$ retrieved reports $\{\mathcal{T}_{\text{ret}}^1, \mathcal{T}_{\text{ret}}^2 \ldots, \mathcal{T}_{\text{ret}}^K\}$ sourced from the training dataset $\mathcal{T}_{\text{final}} = \text{LLM}\left(\mathcal{T}_{\text{init}}, \{\mathcal{T}_{\text{ret}}^1, \mathcal{T}_{\text{ret}}^2 \ldots, \mathcal{T}_{\text{ret}}^K\}\right)$.
The instruction prompt provided to the rewriting agent is meticulously composed of several critical elements (see Prompt~\ref{prompt:rewriting}):

\begin{itemize}[nosep]
\item \textbf{Task description}: A concise overview underscoring the imperative to generate accurate and exhaustive medical reports;
\item \textbf{In-context learning}: Exemplary reference reports to inform and calibrate the rewriting procedure;
\item \textbf{Retrieved-augmentation generation}: Explicit directives for the effective integration of retrieved report content;
\item \textbf{Constraints}: Formalized criteria to ensure medical precision, narrative coherence, and conciseness;
\item \textbf{Chain-of-thought reasoning}: Prompts incentivizing systematic reasoning and critical assessment of the retrieved information prior to report synthesis;
\item \textbf{Multi-agent collaboration}: Deployment of specialized agents dedicated to individual anatomical regions, guaranteeing comprehensive and granular coverage;
\item \textbf{Output format}: Precise specifications delineating the desired structure and scope of the final report, with a mandate to exclude all extraneous details.
\end{itemize}

\begin{prompt}[An example of the instruction prompt for the report rewriting agent]
\label{prompt:rewriting}

\textcolor{dark_green}{// Task Description of Single Anatomical Region:}

Analyze <retrieve region top-k> inputs and compare them with the <AI-generated report> for the same anatomical region. Identify any discrepancies, conflicts, or inconsistencies in findings. Determine if modifications to the <AI-generated report> are necessary to resolve these issues while maintaining clinical fidelity and coherence.

\textcolor{dark_green}{// In-Context Learning:}

<Examples 1 start> The region 0 is abdomen: Bilateral adrenal glands were normal, and no space-occupying lesion was detected. The region 1 is bone: ...  The region 9 is trachea and bronchi: No occlusive pathology was detected in the lumen. Trachea, both main bronchi are open.<Examples 1 end>

\textcolor{dark_green}{// Retrieved-augmentation Generation:}

Please focus on the following medical conditions when refining the AI-generated report: Medical material, Arterial wall calcification, ...
Interlobular septal thickening.

\textcolor{dark_green}{// Constraints:}

Modify the <AI-generated report> only if there is:
   - (a) Clear evidence of a discrepancy, and
   - (b) 90\% or higher confidence that the relevant <retrieve region top-k> data is both accurate and coherent with the report.

\textcolor{dark_green}{// Chain-of-Thought Prompting:}

Split the task into three steps:
Step 1:
For each anatomical region, identify where the discrepancies, conflicts, or inconsistencies are between the <AI-generated report> and the relevant sentences in the <retrieve region top-k> inputs.
Provide a rationale for each criterion.
Step 2:
Create \{region: \{delete: sentence\textsubscript{id}, add: sentence\textsubscript{id}\}\} for each anatomical region based on the analysis in Step 1. If no modifications are necessary for a region, use `0' for both delete and add.
Step 3:
Generate a refined report that ensures clinical accuracy, integrates relevant findings, and conforms to the structure of the <AI-generated report> without deviation.
Step 1 and Step 2 are intermediate reasoning steps and should not be included in the final output.
When you try to answer this question, you need to generate 3-5 different reasoning paths and take the final answer by majority vote among these reasoning paths. Let's think step by step.

\textcolor{dark_green}{// Output Format:}

ONLY OUTPUT THE FINAL REPORT, DO NOT output explanations, comments, or any other non-report content.
<Final Report start>
\end{prompt}

\subsection{Baselines and comparison methods}
\label{sec:sota}

\ourmodel was compared with a matched baseline using the same visual encoder and LLM decoder but without knowledge-augmented retrieval, report refinement, vision–language alignment, or region masking. 

We also compared \ourmodel against several state-of-the-art approaches, including R2GenGPT~\citep{wang2023r2gengpt}, MedVInT~\citep{zhang2023pmcvqa}, RadFM~\citep{wu2025towards}, CT2Rep~\citep{hamamci2024ct2rep}, M3D \citep{bai2024m3d}, and Reg2RG~\citep{chen2025large}. 
R2GenGPT utilizes a transformer architecture to generate radiology reports from images, leveraging a pre-trained GPT backbone. 
MedVInT is a multi-modal transformer that integrates both visual and textual modalities for enhanced report generation. 
Both R2GenGPT and MedVInT are originally tailored for 2D medical imaging tasks.
These two models are adapted in our study to process volumetric chest CT scans by modifying their input layers to accept 3D volume data and adjusting their architectures accordingly.
The remaining four methods, including RadFM, CT2Rep, M3D, and Reg2RG, are specifically designed for generating reports from volumetric medical images. 

\subsection{Evaluation metrics}
\label{sec:metrics}

Report quality was evaluated using established natural language generation metrics, including BLEU (Bilingual Evaluation Understudy)~\citep{papineni2002bleu}, ROUGE (Recall-Oriented Understudy for Gisting Evaluation)~\citep{lin2004rouge}, METEOR (Metric for Evaluation of Translation with Explicit ORdering)~\citep{banerjee2005meteor}, and CIDEr~\citep{vedantam2015cider}. 

Semantic accuracy was assessed using BERTScore~\citep{Zhang2020BERTScore}, which leverages deep contextualized representations derived from pre-trained BERT embeddings.

To appraise the clinical efficacy of the generated narratives, we incorporated a pre-trained BERT model~\citep{yan2022radbert} that classifies the presence or absence of 18 prevalent medical conditions within chest CT reports, including medical material, arterial wall calcification, cardiomegaly, pericardial effusion, coronary artery wall calcification, hiatal hernia, lymphadenopathy, emphysema, atelectasis, lung nodule, lung opacity, pulmonary fibrotic sequelae, pleural effusion, mosaic attenuation pattern, peribronchial thickening, consolidation, bronchiectasis, and interlobular septal thickening. 
The classification performance is summarized by precision, recall, and F1-score for each condition.

We estimated 95\% confidence intervals using 1,000 bootstrap replicates. Paired bootstrap resampling was used for model comparisons, with empirical two-sided $p$ values and a significance threshold of $p < 0.05$.

\subsection{Human evaluation}
\label{sec:human}

Fifty CT cases were randomly selected for expert review. 
Four radiology residents with more than five years of chest imaging experience compared generated reports by either \ourmodel or the baseline method (without KAR, RR, VLA, or RM) with reference reports. 
Reports were presented in random order, and model identity was blinded. 
Each report was rated on a five-point error severity scale: 0, strongly acceptable; 1, acceptable; 2, slightly acceptable; 3, unacceptable; and 4, strongly unacceptable. Inter-rater agreement was measured using Gwet’s AC1 coefficient with quadratic weighting \citep{wongpakaran2013comparison}.
In contrast to Cohen's kappa, Gwet's AC1 is more robust to highly imbalanced rating distributions, as we observed in our evaluation set.

\section*{Data availability}\label{data_availability}
The public RadGenome-ChestCT data used in this study are available at \url{https://huggingface.co/datasets/RadGenome/RadGenome-ChestCT}. The external NYP/WCM cohort contains protected health information and cannot be publicly shared. Access to de-identified derived data may be considered for qualified researchers after institutional review board approval, execution of a data-use agreement, and approval by Weill Cornell Medicine.

\section*{Acknowledgements}\label{acknowledgements}
This work was supported by the U.S. National Science Foundation grant 2145640. The funder had no role in the design and conduct of the study; collection, management, analysis, and interpretation of the data; preparation, review, or approval of the manuscript; and decision to submit the manuscript for publication. We also gratefully acknowledge use of the research computing resources of the Empire AI Consortium, Inc, with support from Empire State Development of the State of New York, the Simons Foundation, and the Secunda Family Foundation.

\section*{CRediT authorship contribution statement}\label{author_contributions}
Study concepts/study design, Y.L., Y.D., Y.P.; manuscript drafting or manuscript revision for important intellectual content, all authors; approval of final version of the submitted manuscript, all authors; agrees to ensure any questions related to the work are appropriately resolved, all authors; literature research, Y.L., Y.P.; experimental studies, all authors; data interpretation and statistical analysis, Y.L., Y.P.; data annotation, E.B., E.T., D.N., H.O., G.S.; and manuscript editing, all authors.

\section*{Competing interests}\label{competing_interests}
The authors declare no competing interests.

\setlength{\bibsep}{3pt plus 0.3ex}
\bibliographystyle{unsrtnat}
\bibliography{references}

\newpage
\appendix
\setcounter{subsection}{0}
\setcounter{table}{0}
\setcounter{figure}{0}
\renewcommand{\thesubsection}{S\arabic{subsection}}
\renewcommand{\tablename}{Supplementary Table}
\renewcommand{\thefigure}{S\arabic{figure}}
\renewcommand{\thetable}{S\arabic{table}}
\captionsetup[figure]{hypcap=false}
\captionsetup[table]{hypcap=false}

\section*{Supplementary Materials}\label{sec:supplementary}

{\begin{table}[!hbpt]
\centering
\caption{Comparison of MonteRET against the baseline model without knowledge-retrieval augmentation and the vision-language alignment strategy.}
\label{tab:exp_retrieval}
\centering
\begin{tabular}{l r@{~(}r@{,}r@{)~~~}r@{~(}r@{,}r@{)~~~} l}
\toprule
& \multicolumn{3}{c}{Baseline} & \multicolumn{3}{c}{MonteRET} & p-value\\
\midrule
\rowcolor[gray]{.9} \multicolumn{8}{l}{\textit{RadGenome-Chest CT dataset}} \\
\midrule
BLEU-1 & 0.452 & 0.445 & 0.459 & \bftab{0.478} & 0.471 & 0.485 & $<0.0001$ \\
BLEU-2 & 0.339 & 0.331 & 0.346 & \bftab{0.370} & 0.362 & 0.377 & $<0.0001$ \\ 
BLEU-3 & 0.267 & 0.259 & 0.275 & \bftab{0.299} & 0.291 & 0.307  & $<0.0001$\\ 
BLEU-4 & 0.219 & 0.211 & 0.226 & \bftab{0.252} & 0.243 & 0.260  & $<0.0001$\\ 
METEOR & 0.413 & 0.406 & 0.420 & \bftab{0.454} & 0.446 & 0.460  & $<0.0001$\\ 
ROUGE-1 & 0.523 & 0.518 & 0.529 & \bftab{0.550} & 0.545 & 0.556  & $<0.0001$\\ 
ROUGE-2 & 0.284 & 0.277 & 0.292 & \bftab{0.321} & 0.313 & 0.329  & $<0.0001$\\ 
ROUGE-L & 0.349 & 0.342 & 0.355 & \bftab{0.379} & 0.372 & 0.386  & $<0.0001$\\ 
CIDEr & \bftab{0.077} & 0.061 & 0.093 & 0.073 & 0.058 & 0.090  & $<0.0001$\\ 
BERTScore & 0.412 & 0.406 & 0.419 & \bftab{0.426} & 0.419 & 0.434  & $<0.0001$\\ 
Precision & 0.376 & 0.344 & 0.413 & \bftab{0.435} & 0.416 & 0.453  & $<0.0001$\\ 
Recall & 0.073 & 0.065 & 0.081 & \bftab{0.406} & 0.387 & 0.425 & $<0.0001$ \\ 
F1 & 0.122 & 0.110 & 0.135 & \bftab{0.420} & 0.403 & 0.436  & $<0.0001$\\ 
\midrule
\rowcolor[gray]{.9} \multicolumn{8}{l}{\textit{External dataset from NYP/WCM}} \\
\midrule
BLEU-1 & 0.217 & 0.202 & 0.230 & \bftab{0.248} & 0.228 & 0.267  & $<0.0001$\\
BLEU-4 & 0.027 & 0.025 & 0.029 & \bftab{0.058} & 0.051 & 0.066  & $<0.0001$\\
METEOR & 0.190 & 0.178 & 0.201 & \bftab{0.220} & 0.205 & 0.235  & $<0.0001$\\
ROUGE-L & 0.134 & 0.128 & 0.140 & \bftab{0.177} & 0.167 & 0.186  & $<0.0001$\\
CIDEr & 0.007 & 0.003 & 0.011 & \bftab{0.015} & 0.006 & 0.025  & $<0.0001$\\
BERTScore & 0.163 & 0.148 & 0.179 & \bftab{0.220} & 0.202 & 0.238  & $<0.0001$\\
Precision & \bftab{0.340} & 0.221 & 0.454 & 0.339 & 0.234 & 0.446  & $>0.05$\\
Recall & \bftab{0.088} & 0.052 & 0.123 & 0.087 & 0.054 & 0.126 & $>0.05$\\
F1 & 0.139 & 0.085 & 0.193 & 0.139 & 0.089 & 0.194 & $>0.05$\\
\bottomrule
\end{tabular}
\end{table}
}
\clearpage
{\begin{table}[!hbpt]
\centering
\caption{Comparison of \ourmodel against several state-of-the-art medical methods on the RadGenome-ChestCT test set: R2GenGPT, MedVInT, RadFM, CT2Rep, M3D, and Reg2RG.}
\label{tab:exp_sota}
\begin{tabular}{l*{7}{c}}
\toprule
{Metrics} & {R2GenGPT} & {MedVInT} & {RadFM} & {CT2Rep} & {M3D} & {Reg2RG} & {MonteRET}\\ 
\midrule
BLEU-1 & 0.433 & 0.443 & 0.442 & 0.444 & 0.436 & 0.473 & \textbf{0.478} \\ 
BLEU-2 & 0.341 & 0.349 & 0.345 & 0.344 & 0.345 & 0.365 & \textbf{0.370} \\ 
BLEU-3 & 0.282 & 0.288 & 0.281 & 0.279 & 0.285 & 0.296 & \textbf{0.299} \\ 
BLEU-4 & 0.242 & 0.246 & 0.237 & 0.236 & 0.245 & 0.249 & \textbf{0.252} \\ 
METEOR & 0.399 & 0.404 & 0.399 & 0.402 & 0.400 & 0.441 & \textbf{0.454} \\ 
ROUGE-L & 0.323 & 0.326 & 0.315 & 0.310 & 0.326 & 0.367 & \textbf{0.379} \\ 
Precision & 0.340 & 0.377 & 0.382 & 0.317 & 0.407 & 0.423 & \textbf{0.435} \\ 
Recall & 0.066 & 0.148 & 0.131 & 0.089 & 0.090 & 0.181 & \textbf{0.406} \\ 
F1 & 0.110 & 0.212 & 0.195 & 0.139 & 0.148 & 0.253 & \textbf{0.420} \\ 
\bottomrule
\end{tabular}
\end{table}
}
\clearpage
{\tablehead{
\toprule
& \multicolumn{3}{c}{\bftab{Baseline}} & \multicolumn{3}{c|}{\bftab{MonteRET}} &
\multicolumn{3}{c}{\bftab{Baseline}} & \multicolumn{3}{c}{\bftab{MonteRET}}\\
\midrule
}
\tabletail{
\midrule
\multicolumn{13}{r}{{Continued on next page}} \\
}
\tablelasttail{
\bottomrule
}
\captionof{table}{Comparison of \ourmodel against the medical report generation model (Baseline) at the region-level.}
\label{tab:exp_region}
\begin{tabular}{l|*{2}{r@{~(}r@{,}r@{)~~}}|*{2}{r@{~(}r@{,}r@{)~}}}
\toprule
& \multicolumn{3}{c}{\bftab{Baseline}} & \multicolumn{3}{c|}{\bftab{MonteRET}} &
\multicolumn{3}{c}{\bftab{Baseline}} & \multicolumn{3}{c}{\bftab{MonteRET}}\\
\midrule
\rowcolor[gray]{.9} & \multicolumn{6}{l|}{Abdomen} & \multicolumn{6}{l}{Bone}\\
BLEU-1  &  0.404  &  0.392  &  0.417  & \bftab{0.414}  &  0.402  &  0.427  &  0.417  &  0.403  &  0.432  & \bftab{0.418}  &  0.403  &  0.433 \\
BLEU-2  &  0.328  &  0.315  &  0.342  & \bftab{0.337}  &  0.324  &  0.351  &  0.346  &  0.329  &  0.362  & \bftab{0.347}  &  0.330  &  0.365 \\
BLEU-3  &  0.286  &  0.272  &  0.300  & \bftab{0.294}  &  0.280  &  0.309  &  0.312  &  0.295  &  0.329  & \bftab{0.313}  &  0.295  &  0.331 \\
BLEU-4  &  0.260  &  0.247  &  0.275  & \bftab{0.267}  &  0.253  &  0.281  &  0.294  &  0.277  &  0.310  & \bftab{0.297}  &  0.279  &  0.314 \\
METEOR  &  0.445  &  0.432  &  0.458  & \bftab{0.461}  &  0.449  &  0.475  &  0.460  &  0.445  &  0.474  & \bftab{0.462}  &  0.447  &  0.477 \\
ROUGE-1  &  0.479  &  0.468  &  0.491  & \bftab{0.488}  &  0.477  &  0.501  &  0.477  &  0.464  &  0.491  & \bftab{0.478}  &  0.465  &  0.493 \\
ROUGE-2  &  0.328  &  0.313  &  0.343  & \bftab{0.335}  &  0.320  &  0.349  &  0.343  &  0.327  &  0.361  & \bftab{0.346}  &  0.329  &  0.361 \\
ROUGE-L  &  0.398  &  0.386  &  0.411  & \bftab{0.400}  &  0.387  &  0.413  &  0.428  &  0.413  &  0.443  & \bftab{0.429}  &  0.414  &  0.443 \\
CIDEr  & \bftab{1.039}  &  0.920  &  1.173  &  1.035  &  0.915  &  1.163  &  1.968  &  1.779  &  2.156  & \bftab{1.989}  &  1.801  &  2.185 \\
BERTScore  &  0.514  &  0.502  &  0.525  & \bftab{0.524}  &  0.512  &  0.534  &  0.562  &  0.549  &  0.573  & \bftab{0.564}  &  0.551  &  0.576 \\
Precision  &  0.507  &  0.454  &  0.562  & \bftab{0.528}  &  0.486  &  0.567  &  0.000  &  0.000  &  0.000  & \bftab{0.197}  &  0.000  &  0.667 \\
Recall  &  0.298  &  0.260  &  0.338  & \bftab{0.659}  &  0.619  &  0.699  &  0.000  &  0.000  &  0.000  & \bftab{0.010}  &  0.000  &  0.032 \\
F1  &  0.375  &  0.337  &  0.414  & \bftab{0.586}  &  0.549  &  0.619  &  0.000  &  0.000  &  0.000  & \bftab{0.019}  &  0.000  &  0.059 \\
\rowcolor[gray]{.9} & \multicolumn{6}{l|}{Breast} & \multicolumn{6}{l}{Heart} \\
BLEU-1  &  0.285  &  0.214  &  0.366  & \bftab{0.305}  &  0.227  &  0.388  &  0.386  &  0.372  &  0.399  & \bftab{0.411}  &  0.398  &  0.426 \\
BLEU-2  &  0.222  &  0.157  &  0.300  & \bftab{0.243}  &  0.169  &  0.332  &  0.314  &  0.301  &  0.328  & \bftab{0.339}  &  0.324  &  0.352 \\
BLEU-3  &  0.197  &  0.130  &  0.271  & \bftab{0.219}  &  0.146  &  0.301  &  0.281  &  0.266  &  0.297  & \bftab{0.303}  &  0.288  &  0.319 \\
BLEU-4  &  0.189  &  0.123  &  0.267  & \bftab{0.210}  &  0.136  &  0.292  &  0.263  &  0.246  &  0.279  & \bftab{0.281}  &  0.266  &  0.296 \\
METEOR  &  0.287  &  0.220  &  0.360  & \bftab{0.306}  &  0.226  &  0.388  &  0.444  &  0.429  &  0.460  & \bftab{0.486}  &  0.471  &  0.499 \\
ROUGE-1  &  0.313  &  0.243  &  0.386  & \bftab{0.334}  &  0.257  &  0.418  &  0.468  &  0.455  &  0.481  & \bftab{0.500}  &  0.486  &  0.513 \\
ROUGE-2  &  0.170  &  0.100  &  0.253  & \bftab{0.191}  &  0.113  &  0.282  &  0.311  &  0.295  &  0.328  & \bftab{0.341}  &  0.324  &  0.357 \\
ROUGE-L  &  0.278  &  0.209  &  0.359  & \bftab{0.299}  &  0.226  &  0.385  &  0.411  &  0.397  &  0.426  & \bftab{0.435}  &  0.420  &  0.449 \\
CIDEr  &  1.213  &  0.489  &  2.124  & \bftab{1.426}  &  0.625  &  2.332  &  1.294  &  1.152  &  1.453  & \bftab{1.385}  &  1.233  &  1.543 \\
BERTScore  &  0.430  &  0.365  &  0.504  & \bftab{0.458}  &  0.386  &  0.533  &  0.532  &  0.519  &  0.544  & \bftab{0.560}  &  0.548  &  0.571 \\
Precision  &  0.114  &  0.000  &  0.300  & \bftab{0.173}  &  0.000  &  0.353  &  0.320  &  0.238  &  0.404  & \bftab{0.530}  &  0.472  &  0.586 \\
Recall  &  0.108  &  0.000  &  0.286  & \bftab{0.157}  &  0.000  &  0.350  &  0.083  &  0.058  &  0.110  & \bftab{0.391}  &  0.342  &  0.443 \\
F1  &  0.110  &  0.000  &  0.286  & \bftab{0.163}  &  0.000  &  0.333  &  0.132  &  0.095  &  0.170  & \bftab{0.450}  &  0.402  &  0.497 \\
\midrule
\multicolumn{13}{r}{{Continued on next page}} \\
\end{tabular}

\begin{tabular}{l|*{2}{r@{~(}r@{,}r@{)~~}}|*{2}{r@{~(}r@{,}r@{)~}}}
\toprule
& \multicolumn{3}{c}{\bftab{Baseline}} & \multicolumn{3}{c|}{\bftab{MonteRET}} &
\multicolumn{3}{c}{\bftab{Baseline}} & \multicolumn{3}{c}{\bftab{MonteRET}}\\
\midrule
\rowcolor[gray]{.9} & \multicolumn{6}{l|}{Esophagus} & \multicolumn{6}{l}{Lung} \\
BLEU-1  & \bftab{0.624}  &  0.607  &  0.640  &  0.622  &  0.605  &  0.637  &  0.282  &  0.271  &  0.292  & \bftab{0.305}  &  0.295  &  0.315 \\
BLEU-2  & \bftab{0.548}  &  0.528  &  0.569  &  0.545  &  0.526  &  0.566  &  0.222  &  0.212  &  0.231  & \bftab{0.240}  &  0.231  &  0.249 \\
BLEU-3  & \bftab{0.516}  &  0.496  &  0.536  &  0.511  &  0.492  &  0.530  &  0.186  &  0.176  &  0.196  & \bftab{0.199}  &  0.189  &  0.208 \\
BLEU-4  & \bftab{0.496}  &  0.477  &  0.516  &  0.488  &  0.468  &  0.506  &  0.162  &  0.151  &  0.171  & \bftab{0.171}  &  0.161  &  0.180 \\
METEOR  &  0.606  &  0.587  &  0.625  & \bftab{0.633}  &  0.616  &  0.650  &  0.326  &  0.316  &  0.337  & \bftab{0.359}  &  0.349  &  0.369 \\
ROUGE-1  &  0.648  &  0.632  &  0.664  & \bftab{0.652}  &  0.636  &  0.667  &  0.385  &  0.377  &  0.395  & \bftab{0.397}  &  0.388  &  0.406 \\
ROUGE-2  &  0.497  &  0.475  &  0.519  & \bftab{0.500}  &  0.478  &  0.520  &  0.229  &  0.219  &  0.238  & \bftab{0.239}  &  0.230  &  0.249 \\
ROUGE-L  & \bftab{0.613}  &  0.595  &  0.632  &  0.605  &  0.589  &  0.622  & \bftab{0.313}  &  0.304  &  0.322  &  0.310  &  0.301  &  0.318 \\
CIDEr  & \bftab{3.234}  &  3.032  &  3.433  &  2.990  &  2.796  &  3.201  & \bftab{0.407}  &  0.326  &  0.487  &  0.400  &  0.320  &  0.487 \\
BERTScore  &  0.736  &  0.722  &  0.748  &  0.736  &  0.725  &  0.748  & \bftab{0.457}  &  0.448  &  0.466  &  0.456  &  0.448  &  0.465 \\
Precision  &  0.284  &  0.059  &  0.546  & \bftab{0.416}  &  0.359  &  0.475  &  0.325  &  0.295  &  0.354  & \bftab{0.376}  &  0.354  &  0.401 \\
Recall  &  0.022  &  0.005  &  0.046  & \bftab{0.628}  &  0.556  &  0.698  &  0.090  &  0.081  &  0.101  & \bftab{0.357}  &  0.332  &  0.381 \\
F1  &  0.040  &  0.009  &  0.084  & \bftab{0.500}  &  0.441  &  0.556  &  0.141  &  0.127  &  0.156  & \bftab{0.366}  &  0.346  &  0.388 \\
\rowcolor[gray]{.9} & \multicolumn{6}{l|}{Mediastinum} & \multicolumn{6}{l}{Pleura} \\
BLEU-1  &  0.364  &  0.355  &  0.372  & \bftab{0.367}  &  0.359  &  0.376  &  0.356  &  0.336  &  0.375  & \bftab{0.366}  &  0.346  &  0.385 \\
BLEU-2  &  0.266  &  0.257  &  0.275  & \bftab{0.270}  &  0.261  &  0.280  &  0.302  &  0.283  &  0.320  & \bftab{0.311}  &  0.292  &  0.329 \\
BLEU-3  &  0.216  &  0.207  &  0.225  & \bftab{0.221}  &  0.211  &  0.230  &  0.274  &  0.254  &  0.293  & \bftab{0.282}  &  0.262  &  0.300 \\
BLEU-4  &  0.186  &  0.177  &  0.196  & \bftab{0.190}  &  0.179  &  0.200  &  0.260  &  0.242  &  0.279  & \bftab{0.268}  &  0.251  &  0.286 \\
METEOR  &  0.412  &  0.401  &  0.422  & \bftab{0.417}  &  0.406  &  0.428  &  0.373  &  0.355  &  0.391  & \bftab{0.384}  &  0.365  &  0.404 \\
ROUGE-1  &  0.422  &  0.414  &  0.430  & \bftab{0.427}  &  0.418  &  0.435  &  0.477  &  0.457  &  0.494  & \bftab{0.490}  &  0.472  &  0.510 \\
ROUGE-2  &  0.230  &  0.220  &  0.241  & \bftab{0.236}  &  0.226  &  0.247  &  0.265  &  0.245  &  0.284  & \bftab{0.272}  &  0.253  &  0.291 \\
ROUGE-L  &  0.326  &  0.318  &  0.336  & \bftab{0.329}  &  0.320  &  0.339  &  0.455  &  0.436  &  0.474  & \bftab{0.466}  &  0.448  &  0.485 \\
CIDEr  &  0.444  &  0.371  &  0.518  & \bftab{0.461}  &  0.392  &  0.537  &  1.517  &  1.327  &  1.706  & \bftab{1.560}  &  1.380  &  1.762 \\
BERTScore  &  0.478  &  0.471  &  0.485  & \bftab{0.482}  &  0.475  &  0.490  &  0.518  &  0.504  &  0.533  & \bftab{0.527}  &  0.513  &  0.543 \\
Precision  &  0.529  &  0.470  &  0.584  & \bftab{0.641}  &  0.593  &  0.687  &  0.132  &  0.079  &  0.194  & \bftab{0.292}  &  0.223  &  0.358 \\
Recall  &  0.234  &  0.205  &  0.264  & \bftab{0.413}  &  0.376  &  0.447  &  0.031  &  0.018  &  0.047  & \bftab{0.098}  &  0.071  &  0.127 \\
F1  &  0.324  &  0.290  &  0.359  & \bftab{0.502}  &  0.466  &  0.535  &  0.051  &  0.029  &  0.075  & \bftab{0.146}  &  0.109  &  0.184 \\
\rowcolor[gray]{.9} & \multicolumn{6}{l|}{Thyroid} & \multicolumn{6}{l}{Trachea and bronchi} \\
BLEU-1  &  0.219  &  0.182  &  0.257  & \bftab{0.224}  &  0.187  &  0.263  & \bftab{0.446}  &  0.430  &  0.462  &  0.445  &  0.428  &  0.460 \\
BLEU-2  &  0.145  &  0.118  &  0.173  & \bftab{0.146}  &  0.121  &  0.175  & \bftab{0.395}  &  0.379  &  0.411  &  0.392  &  0.376  &  0.408 \\
BLEU-3  &  0.114  &  0.094  &  0.136  & \bftab{0.115}  &  0.095  &  0.136  & \bftab{0.363}  &  0.347  &  0.380  &  0.360  &  0.344  &  0.376 \\
BLEU-4  &  0.103  &  0.085  &  0.123  &  0.103  &  0.086  &  0.123  & \bftab{0.341}  &  0.324  &  0.358  &  0.337  &  0.321  &  0.353 \\
METEOR  &  0.225  &  0.182  &  0.269  & \bftab{0.233}  &  0.191  &  0.279  &  0.569  &  0.553  &  0.583  & \bftab{0.572}  &  0.557  &  0.586 \\
ROUGE-1  &  0.257  &  0.204  &  0.307  & \bftab{0.262}  &  0.213  &  0.309  & \bftab{0.570}  &  0.557  &  0.584  &  0.564  &  0.551  &  0.577 \\
ROUGE-2  &  0.099  &  0.069  &  0.133  & \bftab{0.101}  &  0.071  &  0.134  & \bftab{0.429}  &  0.412  &  0.445  &  0.425  &  0.410  &  0.439 \\
ROUGE-L  &  0.216  &  0.170  &  0.262  & \bftab{0.220}  &  0.178  &  0.266  & \bftab{0.530}  &  0.517  &  0.543  &  0.523  &  0.512  &  0.536 \\
CIDEr  &  0.158  &  0.067  &  0.272  & \bftab{0.163}  &  0.070  &  0.271  & \bftab{1.831}  &  1.645  &  2.018  &  1.701  &  1.533  &  1.888 \\
BERTScore  &  0.343  &  0.298  &  0.385  & \bftab{0.350}  &  0.306  &  0.392  & \bftab{0.648}  &  0.638  &  0.659  &  0.643  &  0.633  &  0.653 \\
Precision  &  0.466  &  0.182  &  0.769  & \bftab{0.498}  &  0.231  &  0.769  & \bftab{0.145}  &  0.000  &  0.400  &  0.117  &  0.072  &  0.170 \\
Recall  &  0.200  &  0.067  &  0.355  & \bftab{0.234}  &  0.091  &  0.406  &  0.007  &  0.000  &  0.024  & \bftab{0.164}  &  0.102  &  0.230 \\
F1  &  0.275  &  0.103  &  0.462  & \bftab{0.313}  &  0.140  &  0.490  &  0.013  &  0.000  &  0.044  & \bftab{0.136}  &  0.084  &  0.194 \\
\bottomrule
\end{tabular}
}
\clearpage
{\begin{table}[!hbpt]
\caption{Comparison of \ourmodel against the baseline at the finding-level. MM: Medical Material, AWC: Arterial Wall Calcification, CE: Cardiomegaly, PE: Pericardial Effusion, CAWC: Coronary Artery Wall Calcification, HH: Hiatal Hernia, Lym: Lymphadenopathy, Emp: Emphysema, Ate: Atelectasis, LN: Lung Nodule, LO: Lung Opacity, PFS: Pulmonary Fibrotic Sequela, PLE: Pleural Effusion, MAP: Mosaic Attenuation Pattern, PT: Peribronchial Thickening, Cons: Consolidation, Bro: Bronchiectasis, IST: Interlobular Septal Thickening. Values are reported as mean (95\% CI).}
\label{tab:exp_class}
\resizebox{\textwidth}{!}{
\begin{tabular}{l*{3}{r@{~(}r@{,}r@{)~}}c*{3}{r@{~(}r@{,}r@{)~}}}
\toprule
& \multicolumn{9}{c}{\bftab{Baseline}} && \multicolumn{9}{c}{\bftab{MonteRET}} \\
\cmidrule(r){2-10}\cmidrule(l){12-20}
 &  \multicolumn{3}{c}{\bftab{Precision}}  &  \multicolumn{3}{c}{\bftab{Recall}}  &  \multicolumn{3}{c}{\bftab{F1}}  &&  \multicolumn{3}{c}{\bftab{Precision}}  &  \multicolumn{3}{c}{\bftab{Recall}}  &  \multicolumn{3}{c}{\bftab{F1}} \\
\midrule
MM  &  0.085  &  0.043  &  0.128  &  0.086  &  0.043  &  0.133  &  0.085  &  0.043  &  0.129  && \bftab{0.099}  &  0.053  &  0.150  & \bftab{0.109}  &  0.056  &  0.168  & \bftab{0.103}  &  0.056  &  0.156 \\
AWC  &  0.625  &  0.566  &  0.685  &  0.401  &  0.352  &  0.447  &  0.488  &  0.440  &  0.532  && \bftab{0.635}  &  0.593  &  0.675  & \bftab{0.779}  &  0.737  &  0.819  & \bftab{0.700}  &  0.663  &  0.731 \\
CE  &  0.358  &  0.241  &  0.482  &  0.132  &  0.083  &  0.189  &  0.192  &  0.125  &  0.265  && \bftab{0.654}  &  0.559  &  0.740  & \bftab{0.416}  &  0.338  &  0.489  & \bftab{0.508}  &  0.432  &  0.579 \\
PE  &  0.494  &  0.000  &  1.000  &  0.029  &  0.000  &  0.062  &  0.054  &  0.000  &  0.114  && \bftab{0.690}  &  0.541  &  0.825  & \bftab{0.263}  &  0.183  &  0.345  & \bftab{0.380}  &  0.283  &  0.474 \\
CAWC  &  0.523  &  0.459  &  0.585  &  0.356  &  0.304  &  0.408  &  0.423  &  0.371  &  0.472 && \bftab{0.546}  &  0.501  &  0.589  & \bftab{0.745}  &  0.697  &  0.789  & \bftab{0.630}  &  0.591  &  0.667 \\
HH  &  0.250  &  0.056  &  0.500  &  0.020  &  0.005  &  0.041  &  0.037  &  0.009  &  0.073  && \bftab{0.386}  &  0.334  &  0.437  & \bftab{0.637}  &  0.562  &  0.708  & \bftab{0.480}  &  0.424  &  0.535 \\
Lym  &  0.381  &  0.158  &  0.625  &  0.018  &  0.006  &  0.033  &  0.034  &  0.012  &  0.062  && \bftab{0.526}  &  0.418  &  0.634  & \bftab{0.122}  &  0.090  &  0.159  & \bftab{0.198}  &  0.150  &  0.250 \\
Emp  &  0.342  &  0.270  &  0.417  &  0.179  &  0.138  &  0.227  &  0.235  &  0.186  &  0.289  && \bftab{0.357}  &  0.314  &  0.400  & \bftab{0.511}  &  0.454  &  0.568  & \bftab{0.420}  &  0.376  &  0.462 \\
Ate  &  0.287  &  0.209  &  0.366  &  0.104  &  0.073  &  0.137  &  0.152  &  0.110  &  0.197  && \bftab{0.353}  &  0.305  &  0.401  & \bftab{0.379}  &  0.331  &  0.429  & \bftab{0.365}  &  0.321  &  0.408 \\
LN  &  0.445  &  0.397  &  0.494  &  0.256  &  0.225  &  0.290  &  0.325  &  0.290  &  0.360  && \bftab{0.560}  &  0.519  &  0.603  & \bftab{0.514}  &  0.476  &  0.551  & \bftab{0.536}  &  0.499  &  0.568 \\
LO  &  0.588  &  0.505  &  0.677  &  0.116  &  0.090  &  0.141  &  0.194  &  0.155  &  0.232  && \bftab{0.601}  &  0.538  &  0.667  & \bftab{0.229}  &  0.195  &  0.264  & \bftab{0.331}  &  0.290  &  0.375 \\
PFS  &  0.211  &  0.142  &  0.286  &  0.067  &  0.043  &  0.090  &  0.101  &  0.067  &  0.135  && \bftab{0.417}  &  0.372  &  0.471  & \bftab{0.391}  &  0.344  &  0.438  & \bftab{0.404}  &  0.362  &  0.447 \\
PLE  &  0.614  &  0.375  &  0.857  &  0.066  &  0.032  &  0.107  &  0.119  &  0.057  &  0.186  && \bftab{0.717}  &  0.576  &  0.850  & \bftab{0.168}  &  0.115  &  0.230  & \bftab{0.272}  &  0.194  &  0.353 \\
MAP  &  0.140  &  0.071  &  0.212  &  0.088  &  0.042  &  0.141  &  0.108  &  0.053  &  0.167  && \bftab{0.228}  &  0.168  &  0.284  & \bftab{0.357}  &  0.271  &  0.449  & \bftab{0.278}  &  0.209  &  0.343 \\
PT  & \bftab{0.225}  &  0.000  &  0.571  &  0.013  &  0.000  &  0.033  &  0.024  &  0.000  &  0.061  &&  0.185  &  0.140  &  0.229  & \bftab{0.337}  &  0.268  &  0.412  & \bftab{0.239}  &  0.187  &  0.291 \\
Cons  & \bftab{0.469}  &  0.343  &  0.590  &  0.112  &  0.077  &  0.150  &  0.181  &  0.128  &  0.235  &&  0.405  &  0.326  &  0.486  & \bftab{0.212}  &  0.167  &  0.261  & \bftab{0.278}  &  0.223  &  0.334 \\
Bro  &  0.066  &  0.000  &  0.222  &  0.006  &  0.000  &  0.020  &  0.011  &  0.000  &  0.036  && \bftab{0.217}  &  0.169  &  0.270  & \bftab{0.326}  &  0.258  &  0.392  & \bftab{0.260}  &  0.208  &  0.315 \\
IST  & \bftab{0.276}  &  0.000  &  0.583  &  0.026  &  0.000  &  0.060  &  0.048  &  0.000  &  0.107  &&  0.245  &  0.151  &  0.341  & \bftab{0.156}  &  0.095  &  0.219  & \bftab{0.190}  &  0.116  &  0.261 \\
\bottomrule
\end{tabular}
}
\end{table}

}
\clearpage
{\begin{table}[!hbpt]
\caption{Comparison of different knowledge retrieval methods.}
\label{tab:exp_retrieve}
\footnotesize
\begin{tabular}{l *{4}{r@{~(}r@{,}r@{)~~~~~}}r@{~(}r@{,}r@{)}}
\toprule
 & \multicolumn{3}{c}{No retrieval} & \multicolumn{3}{c}{\makecell[c]{Image-to-image\\retrieval}} & \multicolumn{3}{c}{\makecell[c]{Image-to-text\\retrieval}} & \multicolumn{3}{c}{\makecell[c]{Classification-based\\retrieval}} &  \multicolumn{3}{c}{MonteRET (full)}\\
\midrule
BLEU-1 & 0.456 & 0.428 & 0.486 & 0.457 & 0.425 & 0.490 & 0.462 & 0.432 & 0.492 & 0.499 & 0.471 & 0.524 & \textbf{0.510} & 0.481 & 0.535 \\ 
BLEU-2 & 0.351 & 0.322 & 0.383 & 0.382 & 0.354 & 0.413 & 0.369 & 0.341 & 0.398 & 0.377 & 0.348 & 0.410 & \textbf{0.402} & 0.373 & 0.432 \\
BLEU-3 & 0.283 & 0.251 & 0.318 & 0.308 & 0.278 & 0.342 & 0.295 & 0.267 & 0.323 & 0.307 & 0.275 & 0.340 & \textbf{0.329} & 0.300 & 0.363 \\
BLEU-4 & 0.238 & 0.206 & 0.273 & 0.251 & 0.220 & 0.285 & 0.245 & 0.214 & 0.282 & 0.269 & 0.240 & 0.300 &\textbf{0.279} & 0.247 & 0.310 \\ 
METEOR & 0.406 & 0.378 & 0.435 & 0.418 & 0.389 & 0.447 & 0.413 & 0.386 & 0.441 & 0.452 & 0.429 & 0.476 & \textbf{0.458} & 0.434 & 0.484 \\ 
ROUGE-1 & 0.530 & 0.506 & 0.556 & 0.553 & 0.530 & 0.575 & 0.545 & 0.524 & 0.567 & 0.549 & 0.526 & 0.573 & \textbf{0.565} & 0.543 & 0.586 \\
ROUGE-2 & 0.302 & 0.272 & 0.334 & 0.323 & 0.295 & 0.352 & 0.314 & 0.287 & 0.340 & 0.326 & 0.298 & 0.356 & \textbf{0.342} & 0.312 & 0.372 \\
ROUGE-L & 0.364 & 0.336 & 0.394 & 0.376 & 0.348 & 0.403 & 0.368 & 0.341 & 0.398 & 0.380 & 0.355 & 0.408 & \textbf{0.385} & 0.356 & 0.413 \\ 
CIDEr & 0.102 & 0.040 & 0.205 & 0.104 & 0.038 & 0.197 & 0.117 & 0.048 & 0.216 & 0.084 & 0.045 & 0.134 & \textbf{0.159} & 0.084 & 0.267 \\ 
BERTScore & 0.398 & 0.368 & 0.428 & 0.414 & 0.382 & 0.445 & 0.404 & 0.374 & 0.434 & 0.427 & 0.396 & 0.456 & \textbf{0.435} & 0.407 & 0.463 \\ 
Precision & 0.498 & 0.401 & 0.590 & 0.404 & 0.317 & 0.497 & 0.447 & 0.367 & 0.528 & 0.478 & 0.419 & 0.536 & \textbf{0.528} & 0.470 & 0.590 \\ 
Recall & 0.172 & 0.127 & 0.220 & 0.179 & 0.135 & 0.228 & 0.182 & 0.138 & 0.229 & 0.633 & 0.558 & 0.703 & \textbf{0.659} & 0.594 & 0.724 \\ 
F1 & 0.255 & 0.197 & 0.313 & 0.247 & 0.193 & 0.306 & 0.258 & 0.205 & 0.310 & 0.544 & 0.488 & 0.592 & \textbf{0.585} & 0.538 & 0.633 \\ 
\bottomrule
\end{tabular}
\end{table}
}
\clearpage
{\begin{table}[!hbpt]
\centering
\caption{Comparison of different prompt engineering strategies. CoT: Chain-of-Thought; ICL: In-Context Learning; Cons: Medical Domain Knowledge Constraints; MonteRET: the full model with all prompt strategies.}
\label{tab:exp_prompt}
\resizebox{\textwidth}{!}{
\begin{tabular}{l *{4}{r@{~(}r@{,}r@{)~~~}}r@{~(}r@{,}r@{)}}
\toprule
& \multicolumn{3}{c}{w/o retrieval} & \multicolumn{3}{c}{w/o COT} & \multicolumn{3}{c}{w/o ICL} & \multicolumn{3}{c}{w/o Constraints} & \multicolumn{3}{c}{MonteRET} \\ 
\midrule
BLEU-1 & 0.456 & 0.428 & 0.486 & 0.494 & 0.466 & 0.522 & 0.479 & 0.452 & 0.505 & 0.481 & 0.449 & 0.509 & \bftab{0.509} & 0.483 & 0.535 \\
BLEU-2 & 0.351 & 0.322 & 0.383 & 0.382 & 0.354 & 0.413 & 0.369 & 0.341 & 0.398 & 0.377 & 0.348 & 0.410 & \bftab{0.402} & 0.373 & 0.432 \\
BLEU-3 & 0.283 & 0.251 & 0.318 & 0.308 & 0.278 & 0.342 & 0.295 & 0.267 & 0.323 & 0.307 & 0.275 & 0.340 & \bftab{0.329} & 0.300 & 0.363 \\
BLEU-4 & 0.237 & 0.202 & 0.272 & 0.259 & 0.227 & 0.292 & 0.245 & 0.216 & 0.276 & 0.262 & 0.228 & 0.300 & \bftab{0.279} & 0.248 & 0.311 \\
METEOR & 0.405 & 0.376 & 0.435 & 0.436 & 0.410 & 0.461 & 0.440 & 0.418 & 0.463 & 0.430 & 0.404 & 0.456 & \bftab{0.458} & 0.436 & 0.483 \\
ROUGE-1 & 0.530 & 0.506 & 0.556 & 0.553 & 0.530 & 0.575 & 0.545 & 0.524 & 0.567 & 0.549 & 0.526 & 0.573 & \bftab{0.565} & 0.543 & 0.586 \\
ROUGE-2 & 0.302 & 0.272 & 0.334 & 0.323 & 0.295 & 0.352 & 0.314 & 0.287 & 0.340 & 0.326 & 0.298 & 0.356 & \bftab{0.342} & 0.312 & 0.372 \\
ROUGE-L & 0.364 & 0.336 & 0.394 & 0.370 & 0.342 & 0.398 & 0.361 & 0.338 & 0.385 & 0.378 & 0.350 & 0.406 & \bftab{0.385} & 0.360 & 0.412 \\
CIDEr & 0.100 & 0.040 & 0.193 & 0.153 & 0.070 & 0.270 & 0.092 & 0.029 & 0.193 & 0.154 & 0.070 & 0.273 & \bftab{0.161} & 0.083 & 0.269 \\
BERTScore & 0.399 & 0.368 & 0.430 & 0.419 & 0.389 & 0.448 & 0.403 & 0.375 & 0.428 & 0.424 & 0.395 & 0.455 & \bftab{0.436} & 0.408 & 0.463 \\
Precision & 0.500 & 0.413 & 0.590 & 0.546 & 0.485 & 0.609 & 0.413 & 0.358 & 0.470 & \bftab{0.591} & 0.518 & 0.664 & 0.529 & 0.471 & 0.591 \\
Recall & 0.173 & 0.126 & 0.220 & 0.523 & 0.449 & 0.594 & 0.635 & 0.557 & 0.707 & 0.352 & 0.294 & 0.416 & \bftab{0.661} & 0.594 & 0.726 \\
F1 & 0.257 & 0.195 & 0.315 & 0.533 & 0.481 & 0.584 & 0.500 & 0.447 & 0.550 & 0.440 & 0.385 & 0.498 & \bftab{0.587} & 0.537 & 0.635 \\
\bottomrule
\end{tabular}
}
\end{table}
}
\clearpage
{\begin{table}[!hbpt]
\centering
\caption{Impact of the number of retrieved candidate reports on the performance of the report rewriting agent.}
\label{tab:exp_prompt_size}
\begin{tabular}{l *{3}{r@{~(}r@{,}r@{)~~~}}r@{~(}r@{,}r@{)}}
\toprule
 & \multicolumn{3}{c}{w/o RET} & \multicolumn{3}{c}{TOP-1} & \multicolumn{3}{c}{TOP-3} & \multicolumn{3}{c}{\makecell[c]{TOP-10\\(\ourmodel)}} \\
\midrule
BLEU-1 & 0.456 & 0.428 & 0.486 & 0.442 & 0.418 & 0.464 & 0.467 & 0.438 & 0.495 & \bftab{0.508} & 0.480 & 0.535 \\
BLEU-2 & 0.351 & 0.322 & 0.383 & 0.319 & 0.301 & 0.338 & 0.361 & 0.334 & 0.387 & \bftab{0.401} & 0.372 & 0.429 \\
BLEU-3 & 0.283 & 0.251 & 0.318 & 0.237 & 0.220 & 0.253 & 0.285 & 0.259 & 0.313 & \bftab{0.328} & 0.298 & 0.360 \\
BLEU-4 & 0.237 & 0.202 & 0.272 & 0.182 & 0.168 & 0.197 & 0.233 & 0.208 & 0.260 & \bftab{0.279} & 0.247 & 0.310 \\
METEOR & 0.405 & 0.376 & 0.435 & 0.435 & 0.423 & 0.448 & \bftab{0.461} & 0.440 & 0.482 & 0.459 & 0.436 & 0.485 \\
ROUGE-1 & 0.530 & 0.506 & 0.556 & 0.507 & 0.489 & 0.524 & 0.541 & 0.519 & 0.563 & \bftab{0.565} & 0.543 & 0.588 \\
ROUGE-2 & 0.302 & 0.272 & 0.334 & 0.257 & 0.243 & 0.270 & 0.311 & 0.285 & 0.335 & \bftab{0.341} & 0.312 & 0.371 \\
ROUGE-L & 0.364 & 0.336 & 0.394 & 0.294 & 0.281 & 0.308 & 0.346 & 0.323 & 0.371 & \bftab{0.385} & 0.359 & 0.414 \\
CIDEr & 0.100 & 0.040 & 0.193 & 0.026 & 0.006 & 0.054 & 0.065 & 0.025 & 0.123 & \bftab{0.161} & 0.082 & 0.272 \\
BERTScore & 0.399 & 0.368 & 0.430 & 0.348 & 0.328 & 0.366 & 0.392 & 0.363 & 0.419 & \bftab{0.435} & 0.405 & 0.462 \\
Precision & 0.500 & 0.413 & 0.590 & 0.434 & 0.383 & 0.484 & 0.441 & 0.387 & 0.494 & \bftab{0.530} & 0.473 & 0.587 \\
Recall & 0.173 & 0.126 & 0.220 &\bftab{0.877} & 0.842 & 0.911 & 0.822 & 0.781 & 0.862 & 0.659 & 0.588 & 0.725 \\
F1 & 0.257 & 0.195 & 0.315 & 0.581 & 0.533 & 0.626 & 0.573 & 0.526 & 0.617 & \bftab{0.587} & 0.538 & 0.633 \\
\bottomrule
\end{tabular}
\end{table}
}
\clearpage
{\begin{table}[!hbpt]
\centering
\caption{Ablation study on the key components of MonteRET. KAR: knowledge-augmented retrieval module; RM: region mask strategy; VLA: vision-language alignment strategy; MonteRET: the full model with all components.}
\label{tab:exp_ablation}
\begin{tabular}{l *{3}{r@{~(}r@{,}r@{)~~~}}r@{~(}r@{,}r@{)}}
\toprule
& \multicolumn{3}{c}{w/o KAR} & \multicolumn{3}{c}{w/o RM} & \multicolumn{3}{c}{w/o VLA} & \multicolumn{3}{c}{MonteRET} \\ 
\midrule
BLEU-1 & 0.469 & 0.462 & 0.476 & 0.467 & 0.460 & 0.474 & 0.453 & 0.447 & 0.459 & \bftab{0.478} & 0.471 & 0.485 \\
BLEU-2 & 0.361 & 0.353 & 0.369 & 0.363 & 0.356 & 0.371 & 0.340 & 0.334 & 0.347 & \bftab{0.370} & 0.362 & 0.378 \\
BLEU-3 & 0.292 & 0.283 & 0.300 & 0.294 & 0.286 & 0.303 & 0.266 & 0.260 & 0.273 & \bftab{0.299} & 0.291 & 0.307 \\
BLEU-4 & 0.246 & 0.237 & 0.255 & 0.248 & 0.240 & 0.257 & 0.216 & 0.210 & 0.223 & \bftab{0.251} & 0.243 & 0.259 \\
METEOR & 0.435 & 0.427 & 0.442 & 0.450 & 0.442 & 0.457 & 0.432 & 0.426 & 0.438 & \bftab{0.454} & 0.447 & 0.461 \\
ROUGE-1 & 0.542 & 0.536 & 0.548 & 0.542 & 0.536 & 0.548 & 0.526 & 0.521 & 0.531 & \bftab{0.550} & 0.544 & 0.556 \\
ROUGE-2 & 0.313 & 0.305 & 0.321 & 0.319 & 0.311 & 0.327 & 0.287 & 0.281 & 0.294 & \bftab{0.321} & 0.313 & 0.329 \\
ROUGE-L & 0.379 & 0.372 & 0.387 & 0.367 & 0.359 & 0.374 & 0.341 & 0.335 & 0.347 & \bftab{0.378} & 0.372 & 0.386 \\
CIDEr & \bftab{0.076} & 0.060 & 0.096 & 0.064 & 0.051 & 0.079 & 0.065 & 0.052 & 0.080 & 0.073 & 0.058 & 0.090 \\
BERTScore & 0.426 & 0.418 & 0.433 & 0.413 & 0.406 & 0.420 & 0.398 & 0.392 & 0.404 & \bftab{0.426} & 0.419 & 0.433 \\
Precision & 0.407 & 0.384 & 0.431 & 0.318 & 0.303 & 0.333 & 0.306 & 0.291 & 0.320 & \bftab{0.435} & 0.415 & 0.453 \\
Recall & 0.150 & 0.140 & 0.162 & 0.380 & 0.360 & 0.400 & 0.366 & 0.343 & 0.387 & \bftab{0.406} & 0.387 & 0.425 \\
F1 & 0.219 & 0.205 & 0.234 & 0.346 & 0.332 & 0.362 & 0.333 & 0.318 & 0.349 & \bftab{0.420} & 0.404 & 0.435 \\
\bottomrule
\end{tabular}
\end{table}}
\clearpage
{\begin{table}[!hbpt]
\centering
\caption{Evaluation results from human experts and LLM-as-a-judge. The table shows the empirical cumulative distribution for each report source on Average Error Severity scoring.}
\label{tab:exp_human}
\begin{tabular}{l@{\hspace{20pt}}rccccc}
    \toprule
    & & 0 & $\leq$1 & $\leq$2 & $\leq$3 & $\leq$4 \\
    \midrule
    \multicolumn{2}{l}{Expert I} \\
    & Baseline & 0.3696 & 0.6087 & 0.6304 & 0.6957 & 1.0000 \\ 
    & MonteRET & 0.3043 & 0.4348 & 0.5870 & 0.6087 & 1.0000 \\ 
    \multicolumn{2}{l}{Expert II}  \\
    & Baseline & 0.6087 & 0.7826 & 0.8696 & 0.9783 & 1.0000 \\ 
    & MonteRET & 0.2609 & 0.3913 & 0.5870 & 0.8478 & 1.0000 \\ 
    \multicolumn{2}{l}{Expert III}  \\
    & Baseline & 0.4565 & 0.7609 & 0.8913 & 0.9565 & 1.0000 \\ 
    & MonteRET & 0.3696 & 0.4783 & 0.6304 & 0.7609 & 1.0000 \\ 
    \multicolumn{2}{l}{Expert IV}  \\
    & Baseline & 0.4348 & 0.7391 & 0.7609 & 0.8913 & 1.0000 \\ 
    & MonteRET & 0.2609 & 0.5435 & 0.6304 & 0.7391 & 1.0000 \\ 
    \multicolumn{2}{l}{LLM}  \\
    & Baseline & 0.1304 & 0.3478 & 0.7609 & 0.9348 & 1.0000 \\ 
    & MonteRET & 0.0435 & 0.1739 & 0.4130 & 0.6957 & 1.0000 \\ 
    \bottomrule
\end{tabular}
\end{table}
}
\clearpage

\begin{figure}[!htbp]
    \centering
    \includegraphics[width=.6\textwidth]{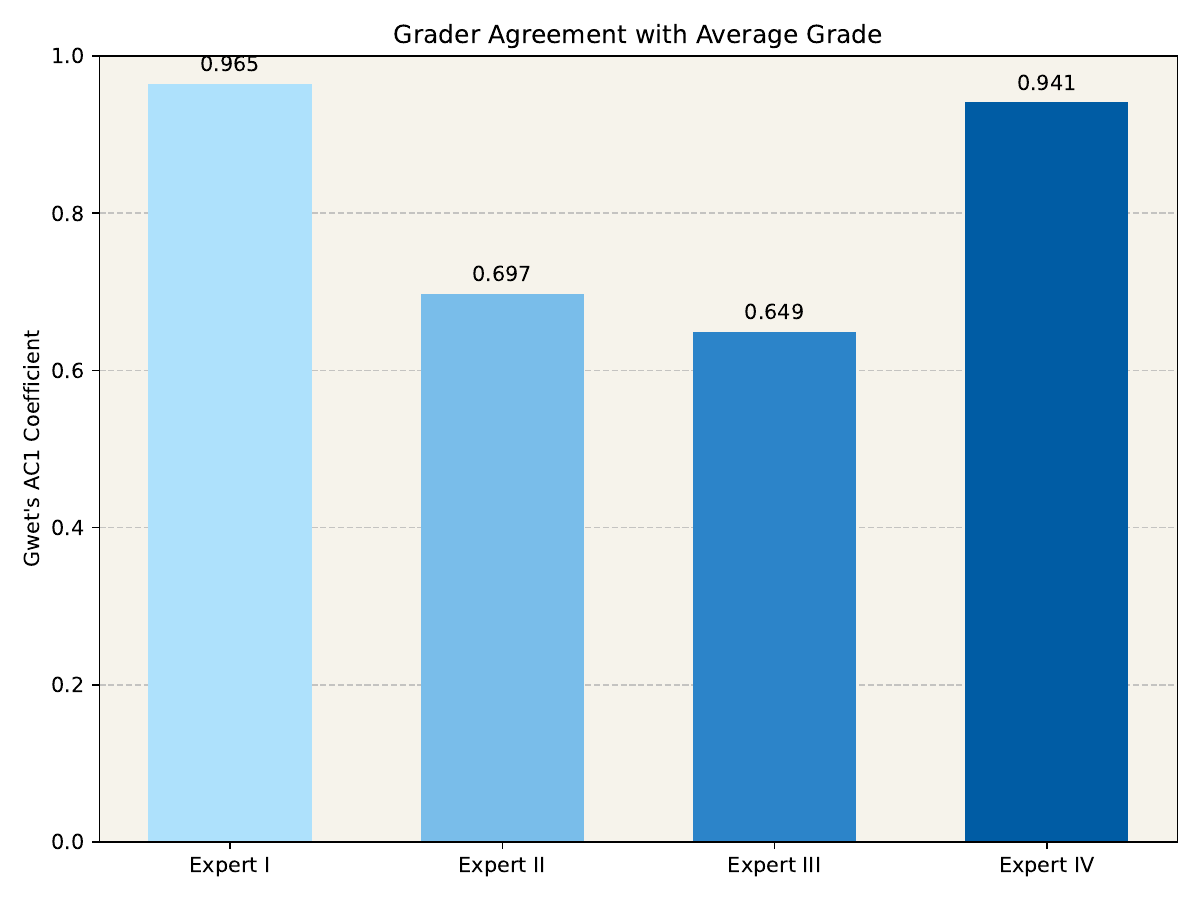}
    \caption{The inter-rater agreement among four radiologists, measured by Gwet's AC1 coefficient with respect to the result of majority voting.}
    \label{fig:exp_human_kappa}
\end{figure}

\newpage
\clearpage

\subsection{Experimental details of human expert evaluation}
\label{sec:sm_human_eval}
\paragraph{Overall Instructions.}
You will be presented with a reference CT report and an AI-generated report.
Each report consists of ten regions, each corresponding to a specific part of the medical report.
For each region, you are required to score the error severity in the report, using the provided scoring guidelines.
The AI-generated report is produced by different models, which are randomly ordered to avoid bias.

\paragraph{Scoring Guidelines}
\begin{enumerate}[start=0,nosep]
    \item \textbf{Strongly Acceptable}: The report accurately describes the absence of abnormalities or the presence of normal findings, consistent with the reference report.
    \item \textbf{Acceptable}: The report contains minor inaccuracies or omissions that do not impact patient care, such as slight discrepancies in measurements or descriptions of non-critical findings.
    \item \textbf{Slightly Acceptable}: The report contains errors that could lead to non-urgent clinical actions or follow-ups, such as misidentification of a condition or omission of a significant finding that requires monitoring.
    \item \textbf{Unacceptable}: The report contains errors that could lead to urgent clinical actions or interventions, such as incorrect identification of a condition that requires prompt treatment.
    \item \textbf{Strongly Unacceptable}: The report contains critical errors that could lead to immediate and potentially life-threatening clinical actions, such as misdiagnosis of a severe condition or omission of a critical finding that requires emergency intervention.
\end{enumerate}

\subsection{Training Configuration and Hyperparameters}
\label{sec:sm_training_config}
For the training of report generation model, we set the learning rate to 5e-5 and used a batch size of 2 per device with gradient accumulation over 2 steps, resulting in an effective batch size of 4. The model was trained for 10 epochs using the constant learning rate scheduler with 20 warmup steps. The vision-language alignment module was trained with a per-device batch size of 16, gradient accumulation over 8 steps (effective batch size 128), and 10 epochs. All experiments were conducted using bfloat16 precision. 
The LoRA configuration is set with a rank of 8 and a scaling factor of 32, which empirically balances the trade-off between model expressiveness and computational efficiency.
The multi-class classification model for knowledge-augmented retrieval was trained with a learning rate of 3e-4, a batch size of 128 per device, and gradient accumulation over 8 steps (effective batch size 1024). 
The segmentation model was adopted from the pretrained SAT-Pro model, which was configured with 256 object queries.
The experiments were implemented using the Hugging Face Transformers library (v4.28.1) and PyTorch framework (v2.5.0) with Python 3.10.0 as the programming language.
All training and inference experiments were conducted on an NVIDIA A100 GPU with 80GB of memory, and the training time for the report generation model was approximately 40 hours, while the vision-language alignment module took around 24 hours to train.

\end{document}